%% file: main.tex
\documentclass[11pt]{article}

\usepackage{acl}

\usepackage{times}
\usepackage{latexsym}
\usepackage[T1]{fontenc}
\usepackage[utf8]{inputenc}
\usepackage{microtype}
\usepackage{inconsolata}
\usepackage{graphicx}
\usepackage{subcaption}  

\usepackage{amsmath}
\usepackage{amssymb}
\usepackage{booktabs}
\usepackage{array}      
\usepackage{multirow}
\usepackage{pifont}
\usepackage{xcolor}
\usepackage{colortbl}
\usepackage{pifont}
\usepackage{wasysym}
\usepackage{booktabs}
\usepackage{makecell}
\usepackage{hyperref}

\usepackage{listings}
\lstdefinestyle{promptstyle}{
  basicstyle=\ttfamily\scriptsize,
  breaklines=true,
  breakindent=0pt,
  postbreak=\mbox{\textcolor{gray}{$\hookrightarrow$}\space},
  columns=fullflexible,
  keepspaces=true,
  frame=single,
  framesep=4pt,
  rulecolor=\color{gray!50},
  xleftmargin=6pt,
  xrightmargin=2pt,
  aboveskip=6pt,
  belowskip=6pt,
}
\usepackage{xspace}
\newcommand{\awtn}{AppWorld$_\text{TN}$\xspace}
\newcommand{\awtc}{AppWorld$_\text{TC}$\xspace}

\definecolor{deltagood}{RGB}{0,140,60}
\definecolor{deltabad}{RGB}{200,0,0}
\definecolor{ourshl}{RGB}{238,249,238}
\definecolor{verifhl}{RGB}{253,242,230}

\definecolor{blockhl}{RGB}{236,236,236}

\title{DRACO: Fine-Grained Credit Assignment with Dynamic Rubrics\\
  for Long-Horizon Agent Training}

\author{%
  Shubham Gandhi$^{1,*}$ \quad Saurabh Goyal$^{2,*}$ \quad
  Kiran Kate$^{2}$ \quad Yara Rizk$^{2}$ \\[0.5em]
  \normalsize $^{1}$Carnegie Mellon University \qquad $^{2}$IBM Research\\
  \normalsize Correspondence: \texttt{srgandhi@andrew.cmu.edu}
}
\date{}

\begin{document}
\maketitle

\renewcommand{\thefootnote}{\fnsymbol{footnote}}
\footnotetext[1]{Equal contribution.}
\renewcommand{\thefootnote}{\arabic{footnote}}




\begin{abstract}

Reinforcement learning from verifiable rewards works well when a task has a programmatic checker, but most long-horizon agent domains have none. We work in the outcome-blind setting, where ground-truth success signals are not available. Multi-criteria rubrics are a popular way to supply such a reward; they are scored once per trajectory, but a single scalar is a poor signal across tens of steps. We propose \textbf{DRACO}: Distributing Rubric-based Advantage for Credit Optimization. It generates rubrics dynamically during training to track the policy's evolving capability, scores those rubrics once per completed trajectory, and redistributes that judgment over the steps responsible for annotated rubrics to produce differentiated per-step advantages in GRPO. The redistribution is closed-form and does not introduce any trained attribution module. On AppWorld, DRACO gains 15.9 points over the base model and 5.3 points over GRPO trained with a sparse ground-truth reward, despite not using any verifiers itself. On out-of-domain $\tau$-bench, it gains 5.3 points over the base model even without a frontier judge, beating both ground-truth-reward training and other rubric-based training settings.\footnote{The code for DRACO is available at \href{https://github.com/IBM/draco}{https://github.com/IBM/draco}.}

\end{abstract}

\input{sections/01-introduction}

\input{sections/02-related-work}

\input{sections/03-method}
\input{sections/04-experimental-setup}

\input{sections/06-results}

\input{sections/08-conclusion}

\section{Limitations}
\input{sections/09-limitations}

\section{Ethics Statement}
\input{sections/10-ethics}

\bibliography{references}

\appendix
\input{sections/11-appendix}

\end{document}

%% file: sections/01-introduction.tex
\section{Introduction}
\label{sec:intro}

Reinforcement learning from verifiable rewards (RLVR) has driven much of the recent progress in large language models (LLMs), from mathematical reasoning \citep{deepseekR1,grpo} to interactive tool-using agents \citep{appworld, webarena, taubench}.
The recipe sidesteps a learned reward model: a programmatic verifier (unit tests for code, exact-match for math, task-pass checks for agents) supplies a reliable terminal reward that is difficult to game.
Yet it rests on an assumption that fails in many real-world settings: that a ground-truth verifier exists.
Many agent domains, such as customer-support and open-ended research agents, have no programmatic oracle for success, and constructing one is often as hard as the task itself.


We study this problem under an \emph{outcome-blind} regime: the training signal is derived entirely from process criteria, with no access to a ground-truth success or gold-answer signal at any point during training.
This is a strictly harder setting than most of the reward literature assumes, and it interacts sharply with a second difficulty specific to \emph{long-horizon} agents.
A trajectory on a benchmark such as AppWorld \citep{appworld} spans tens of interdependent tool-calling steps.
Even when a reward is available, attributing a single trajectory-level scalar uniformly to every step is statistically wasteful and can be actively harmful: successful trajectories contain redundant or lucky steps, and failed trajectories contain mostly correct ones \citep{rudder,hindsightCredit}.
This is the classic credit-assignment problem, and it is well studied for agents when a terminal reward exists \citep{salt,gigpo,sparl}.
What is missing is credit assignment in the outcome-blind regime: routing a rubric-derived signal, rather than a verifier outcome, to individual steps.

\begin{figure*}[!t]
    \centering
    \includegraphics[width=\textwidth]{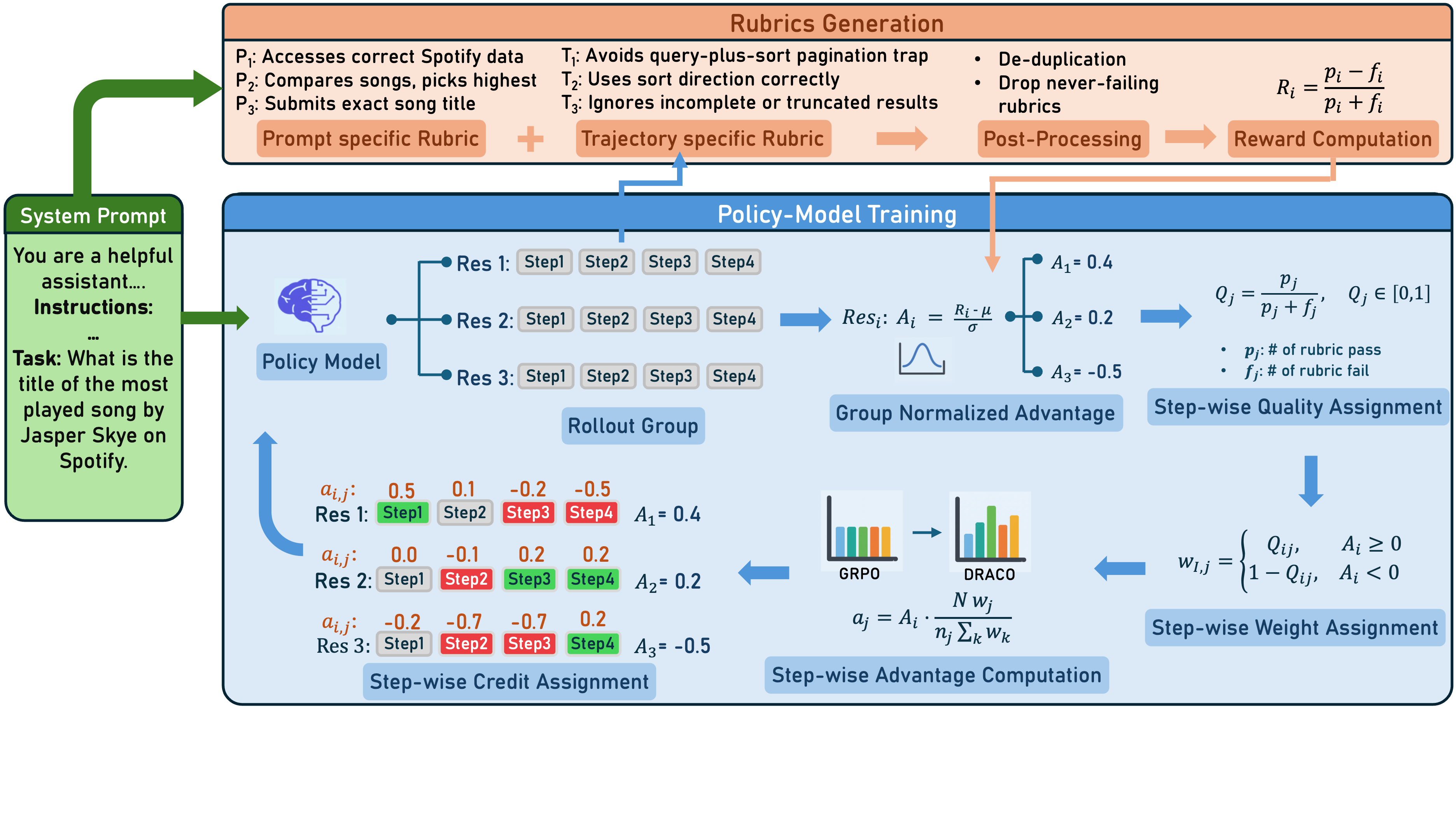}
    \vspace{-48pt}
    \caption{Overview of DRACO. Top: for each task, the judge proposes criteria from the instruction and each sampled trajectory; proposals are merged, deduplicated, and scored to produce an outcome-blind reward $R_i$ (Section 3.2). Bottom: within a rollout group, GRPO normalizes rewards into trajectory advantages $A_i$; DRACO then reallocates each $A_i$ across steps according to the judge's per-criterion verdicts, yielding step advantages $a_j$ that concentrate credit on the steps the rubric implicates (Section 3.3).}
    \label{fig:image}
\end{figure*}

These two difficulties define our target, and DRACO addresses them with two complementary components.
First, a fixed rubric authored once for a task distribution cannot anticipate the diverse ways a long trajectory can succeed or fail; we therefore generate \textbf{dynamic, per-trajectory rubrics} that adapt the evaluation criteria to what a given rollout actually does.
Second, to convert rubric verdicts into a learning signal that respects the structure of a long trajectory, we perform \textbf{distribution of rubric-conditioned advantage to steps}, mapping criterion-level verdicts to per-step contributions that are then used within GRPO \citep{grpo}. Figure~\ref{fig:image} shows an overview of the two steps.
Together, these instantiate the two axes (task-adaptive coverage and faithful per-step attribution) that the ``verification horizon'' framing \citep{verificationhorizon} argues are jointly necessary when verification is harder than generation.
Crucially, neither component assumes access to a ground-truth outcome at training time, unlike prior methods which tie step rewards to terminal verifiers or gold answers \citep{arco, scribe, rtt}.


We instantiate DRACO on AppWorld, a benchmark whose every previously reported RL result (PPO, GRPO, RLOO, DPO variants, and LOOP \citep{loop}) trains against the environment's ground-truth unit tests.
To our knowledge, ours is the first RL training on AppWorld that never accesses that signal.
On \awtn, DRACO achieves 85.3 TGC ($p^1$), a +15.9-point gain over the untrained policy, with zero-shot transfer to $\tau$-bench and performance at or above a verifier-trained run of the same budget.

Our contributions include: 
  (1) formulating \emph{outcome-blind} rubric-based RL for long-horizon tool-using agents, positioning it against the rubric, step-credit, and process-reward literature, nearly all of which anchor their signal to a ground-truth outcome (Section~\ref{sec:related}); 
%
  (2) proposing two complementary mechanisms: dynamic per-trajectory rubric generation, and distribution of rubric-conditioned advantage to steps within GRPO, neither of which requires a verifier (Section~\ref{sec:method}); 
  the credit rule satisfies seven formal properties including total-push conservation, sign preservation, and length independence (Appendix~\ref{sec:appendix:credit:properties});
  %
  and, (3) reporting results on AppWorld and $\tau$-bench, and analyze the two components separately and jointly, an outcome-reward baseline, a self-judge in place of the frontier judge, and how training affects efficiency and the reward the policy model optimizes (Section~\ref{sec:exp}).
    

%% file: sections/02-related-work.tex
\section{Related Work}
\label{sec:related}

Training long-horizon agents without a verifier raises two coupled problems: what to reward when task success cannot be checked, and how to attribute that reward to individual steps. Table~\ref{tab:comparison} positions prior work along these axes.

\paragraph{Rubric-based rewards.}
Several methods evolve a rubric during training (adding, pruning, or rewriting criteria as the policy improves) and score it once on the finished rollout~\cite{drtulu,evorubric,onlinerubrics,rubricarm}.
A second family scores a fresh rubric at every position, binding criteria to a task-specific decomposition: per-step generation~\cite{arco}, deep-research stages~\cite{rubricem}, or subgoal prototypes~\cite{scribe}.
This is expensive (a judge call per position) and fixes the notion of a step to the task at hand.
\citet{rtt} is the only method that propagates a trajectory-level rubric score to individual tokens, by training a learned discriminator based on fixed criteria.
DRACO retains the evolving, trajectory-scored rubric of the first family but redistributes the judgment over steps in closed form, requiring no per-position judge call and no learned module.

\paragraph{Step-level credit assignment.}
Most fine-grained credit for LLM agents decomposes the episode outcome: via trained progress estimators~\cite{sparl,sweetrl}, recurrence across rollouts~\cite{salt,gigpo}, or a reference model's likelihood of the gold answer~\cite{trace}.
All require a reliable outcome signal.
A second group attributes credit post hoc with an LLM: hindsight $Q$-value refinement~\cite{hcapo}, failure-step localization~\cite{hintsd}, or per-action good/bad labels~\cite{agentevolver}.
A third specializes to an action space: tool-call formats~\cite{steptool}, CLI commands~\cite{cliagents}, or next-state signals that bypass the episode outcome~\cite{openclawrl}.
DRACO's attribution departs on three counts: it consumes no ground-truth outcome, it assumes no particular action structure, and the redistribution is closed-form with no learned component.

\newcommand{\cmark}{\ding{51}}
\newcommand{\xmark}{\ding{55}}
\newcommand{\pmark}{\LEFTcircle}

\begin{table*}[t]
\centering
\small
\setlength{\tabcolsep}{6pt}
\resizebox{\textwidth}{!}{%
\begin{tabular}{lcccccc}
\toprule
Method
& \makecell{Outcome-\\Blind}
& \makecell{Rubric\\Reward}
& \makecell{Dynamic\\Rubrics}
& \makecell{Traj.-Level\\Scoring}
& \makecell{Step\\Attribution}
& \makecell{No Learned\\Attribution} \\
\midrule
DR T\"ulu~\cite{drtulu} & \cmark & \cmark & \cmark & \cmark & \xmark & -- \\
EvoRubric~\cite{evorubric} & \cmark & \cmark & \cmark & \cmark & \xmark & -- \\
Online Rubrics Elicit.~\cite{onlinerubrics} & \cmark & \cmark & \cmark & \cmark & \xmark & -- \\
RubricARM~\cite{rubricarm} & \xmark & \cmark & \cmark & \cmark & \xmark & -- \\
\midrule
ARCO~\cite{arco} & \pmark & \cmark & \cmark & \xmark & \cmark & \xmark \\
RubricEM~\cite{rubricem} & \cmark & \cmark & \cmark & \xmark & \cmark & \cmark \\
SCRIBE~\cite{scribe} & \pmark & \pmark & \cmark & \xmark & \xmark & \xmark \\
\midrule
Rubrics to Tokens~\cite{rtt} & \pmark & \cmark & \xmark & \cmark & \cmark & \xmark \\
AgentEvolver~\cite{agentevolver} & \xmark & \pmark & \xmark & \cmark & \cmark & \cmark \\
HCAPO~\cite{hcapo} & \xmark & \xmark & \xmark & \cmark & \cmark & \cmark \\
SALT~\cite{salt} & \xmark & \xmark & \xmark & \cmark & \cmark & \cmark \\
\midrule
\textbf{DRACO (ours)} & \cmark & \cmark & \cmark & \cmark & \cmark & \cmark \\
\bottomrule
\end{tabular}}
\caption{Comparison with the closest prior work. \cmark~yes, \xmark~no, \pmark~partial,
--~N/A. \emph{Outcome-Blind}: the training reward consults no ground-truth
outcome (\pmark: judge plus verifier). \emph{Rubric Reward}: reward derived
from written criteria (\pmark: a criteria-guided judge emitting one verdict rather than
per-criterion scores). \emph{Dynamic Rubrics}: criteria that change
during training, driven by the training algorithm rather than generated per
instance by a fixed procedure. \emph{Traj.-Level Scoring}: the rubric is scored
once on the finished rollout, not at each position. \emph{Step
Attribution}: differentiated credit reaches the policy update at step
granularity. \emph{No Learned Attribution}: attribution requires
no trained module.}
\label{tab:comparison}
\end{table*}

%% file: sections/03-method.tex
\section{Method}
\label{sec:method}

A policy model $\pi_\theta$ interacts with an environment over a long-horizon episode.
Given a task instruction $x$, it produces a trajectory
$\tau = (s_1, a_1, \dots, s_T, a_T)$ of interleaved reasoning and tool calls,
with $T$ on the order of tens of steps. We assume no verifier of task success at
training time; the reward must come from process criteria alone. Our method
has two parts: a dynamic per-trajectory rubric scored by a frozen judge
(Section~\ref{sec:method:rubric}), and a step-level credit rule that turns the
judge's verdicts into per-step advantages for GRPO
(Section~\ref{sec:method:credit}).

\subsection{Background: GRPO}
\label{sec:method:grpo}

We train with Group Relative Policy Optimization \citep{grpo, deepseekR1}. For
each task $x$, we sample a group of $G$ trajectories $\{\tau_i\}_{i=1}^{G}$ and
give each a scalar reward $R_i$. GRPO standardizes rewards within the group to
form the advantage
\begin{equation}
  A_i = \frac{R_i - \mathrm{mean}(\{R_j\}_{j=1}^{G})}{\mathrm{std}(\{R_j\}_{j=1}^{G})}.
  \label{eq:grpo-adv}
\end{equation}
Let trajectory $\tau_i$ emit tokens $y_{i,1}, \dots, y_{i,N_i}$. GRPO assigns the
same scalar $A_i$ to every one of these tokens and takes a policy-gradient step in
which each token's log-probability is pushed by its advantage,
\begin{equation}
  \nabla_\theta \mathcal{J}
  = \mathbb{E}\!\left[\, \sum_{t=1}^{N_i} A_i \,
    \nabla_\theta \log \pi_\theta(y_{i,t} \mid y_{i,<t}, x) \right].
  \label{eq:grpo-pg}
\end{equation}
The advantage is thus a per-token multiplier on the log-probability gradient:
$A_i > 0$ makes the tokens of $\tau_i$ more likely in their context, $A_i < 0$
less likely. Because a single $A_i$ multiplies all $N_i$ tokens, every decision in
the episode receives identical credit. In RLVR, $R_i$ is a terminal verifier
outcome; we change both the source of $R_i$ (a rubric judge, not a verifier;
Section~\ref{sec:method:rubric}) and its uniform use, replacing $A_i$ with a
per-step advantage $a_j$ that concentrates on the steps the judge implicates
(Section~\ref{sec:method:credit}).

\subsection{Dynamic Per-Trajectory Rubrics}
\label{sec:method:rubric}

Instead of building one generic rubric for the entire task distribution, we build one rubric per
task and score it per trajectory. As shown in figure~\ref{fig:image}, a judge first proposes criteria from the task
instruction alone, then extends them once per sampled rollout, adding the sub-goals that
rollout reveals and the ways the policy model actually fails; the proposals from all $G$
rollouts are merged into a single set $\mathcal{R}=\{c_1,\dots,c_K\}$ with duplicates
removed. All generation calls
are instructed to keep criteria \emph{mutually exclusive and collectively exhaustive}~\citep{highrewardtail}, since the reward in Eq.~\eqref{eq:rubric-reward} is a rate and two overlapping
criteria would count one mistake twice. We then apply \emph{discriminative dropout},
keeping a criterion only if some group member failed it. A frozen external judge scores each trajectory against every surviving criterion,
returning \emph{pass}, \emph{fail}, or \emph{not applicable}, coarse by
design~\citep{discretizingrm}, together with a justification and the steps responsible,
which the credit rule consumes (Section~\ref{sec:method:credit}). Letting $p_i$ and $f_i$
count the applicable passes and fails,
\begin{equation}
  R_i = \frac{p_i - f_i}{p_i + f_i},
  \label{eq:rubric-reward}
\end{equation}
with $R_i = 0$ when no criterion applies. The criterion set is shared across the group, but applicability is not: a criterion about
pagination is moot for a trajectory that never listed a collection. Because $R_i$ normalizes by the verdicts actually cast rather than by $K$, trajectories
with different numbers of applicable criteria remain comparable within the group.
Note that the reward is completely outcome-blind as no task-success or gold-answer term appears.

\subsection{Rubric-Conditioned Step Credit}
\label{sec:method:credit}

Placing $R_i$ into Eq.~\eqref{eq:grpo-adv} still gives every token the same
advantage $A_i$. But each criterion is usually decided by a few steps, not the
whole trajectory. Our second change \emph{reallocates} $A_i$ across the steps of
$\tau_i$ according to which steps the judge's verdicts implicate, without
changing the trajectory's total push. We index the steps of $\tau_i$ by
$j = 1, \dots, M$ (one step is one agent turn, i.e.\ one emitted code block).
Each response token belongs to exactly one step, or to a \emph{gap} (turn glue,
tool-result echoes) that belongs to no step; gap tokens receive no credit and are
excluded from the reallocation.

\paragraph{Step quality.}
When evaluating against the rubric, the judge cites the steps that justify each criterion's rating. Let $p_j$ and $f_j$ be the number of passed and failed criteria
that cite step $j$. The quality of step $j$ is the pass fraction of the criteria
citing it,
\begin{equation}
  Q_j = \frac{p_j}{p_j + f_j}, \qquad Q_j \in [0,1].
  \label{eq:stepq}
\end{equation}
A step cited by no criterion inherits $\bar{Q}$, the mean $Q$ over cited steps.

\paragraph{Sign-preserving step weights.}
The sign of $A_i$ fixes whether the whole trajectory is reinforced
($A_i \ge 0$, a ``winner'') or suppressed ($A_i < 0$, a ``loser''); credit only
decides where within the trajectory that push lands. The step weight makes the
reallocation sign-correct:
\begin{equation}
  w_j =
  \begin{cases}
    Q_j, & A_i \ge 0 \ \text{(reinforce good steps)},\\[2pt]
    1 - Q_j, & A_i < 0 \ \text{(suppress bad steps)}.
  \end{cases}
  \label{eq:stepw}
\end{equation}
A step all of whose citing criteria agree takes weight $0$ and is left out of the
update, which is the intended reading: on a winner, a step every criterion failed
should not be reinforced.

\paragraph{Step advantage.}
Credit is assigned at the step level: every token in step $j$ receives the same
advantage $a_j$, since the rubric grades steps, not tokens. Let $n_j$ be the
number of tokens in step $j$ and $N = \sum_k n_k$ the tokens lying inside some
step. The step advantage is
\begin{equation}
  a_j = A_i \cdot \frac{N\, w_j}{n_j \sum_k w_k},
  \label{eq:credit}
\end{equation}
and it replaces the uniform $A_i$ on every token of step $j$ in the GRPO update.
What the rule equalizes is each step's \emph{total} contribution: step $j$
contributes $n_j a_j = A_i N w_j / \sum_k w_k$, which depends on its quality weight
and not on its length. A step therefore earns influence by being judged good, not
by being verbose, and the $1/n_j$ factor is what spreads that fixed total over
however many tokens the step happens to contain. Summing over steps conserves the
trajectory's total push,
\begin{equation}
  \sum_j n_j\, a_j = A_i N \frac{\sum_j w_j}{\sum_k w_k} = A_i N,
  \label{eq:conserve}
\end{equation}
which is exactly the total baseline GRPO applies to those same $N$ tokens. Credit
reallocation therefore never inflates or deflates a trajectory's overall influence;
it moves the existing influence onto the steps that earned it. Because every
$w_j \ge 0$, no $a_j$ ever takes the opposite sign to $A_i$, so credit is never
inverted relative to the judge's trajectory-level verdict. When the rubric cites no
step at all there is nothing to reallocate on, and the update falls back to
baseline GRPO. A full derivation, worked example, a traced real rollout, and the
configuration are given in Appendix~\ref{sec:appendix:credit}.

\paragraph{Training loop.}
In each step: sample a group of $G$ trajectories for each task; generate the rubric
in three stages, merging into one set shared by the group; score every
trajectory against that set with the frozen judge, obtaining a verdict and its
step citations per criterion; drop the criteria no member failed and compute
Eq.~\eqref{eq:rubric-reward} over the survivors; standardize within the group
into $A_i$ (Eq.~\eqref{eq:grpo-adv}); compute step qualities
Eq.~\eqref{eq:stepq}, weights Eq.~\eqref{eq:stepw}, and step advantages
Eq.~\eqref{eq:credit} from the retained criteria's citations; update the policy model.
The four settings in Section~\ref{sec:setup:settings} isolate these parts by removing
per-trajectory rubrics, step credit, or both from DRACO.

%% file: sections/04-experimental-setup.tex
\section{Experimentation}
\label{sec:exp}

\subsection{Experimental Setup}
\label{sec:setup}

\subsubsection{Settings}
\label{sec:setup:settings}

We evaluate four outcome-blind settings initialized from the same base policy model (\texttt{Qwen3.6-27B}), each named for what it removes from DRACO: (i) \textbf{DRACO w/o Dyn.\ \& Cred.} (static rubric and no step credit), (ii) \textbf{DRACO w/o Dyn.} (static rubric with rubric-conditioned step credit; Section~\ref{sec:method:credit}), (iii) \textbf{DRACO w/o Cred.} (dynamic rubrics without step credit), and (iv) \textbf{DRACO}. An \textbf{outcome reward model} trained on AppWorld unit tests (binary) serves as an outcome-aware reference.

\subsubsection{Training and Implementation}
\label{sec:setup:training}

\paragraph{Data and models.}
We train on the AppWorld training split (90 tasks). We report results on \texttt{Qwen3.6-27B} (used for all ablations) and \texttt{Qwen2.5-32B-Instruct} \citep{qwen25}.

\paragraph{RL and hyperparameters.}
All runs train LoRA adapters \citep{lora} using GRPO (Section~\ref{sec:method:grpo}). Each training step uses batch size $B=16$ with GRPO group size $G=6$ per task. All settings share a single pre-committed hyperparameter configuration, differing strictly in their reward signals (Section~\ref{sec:setup:settings}), and every run trains on 8 H100 GPUs.
All training-time reward operations such as rubric generation, union, scoring, and credit reallocation use a single judge model (\texttt{GPT-5.4}, temperature $0.1$) for consistency.  Appendix~\ref{sec:appendix:rubrics} details static rubric construction and lists the resulting criteria (Table~\ref{tab:staticrubrics}), Appendix~\ref{sec:appendix:hparams} gives full hyperparameters (Table~\ref{tab:hparams}) and model-specific serving configurations, and Appendix~\ref{sec:appendix:prompts} includes every judge prompt and the agent's system prompt.

\subsubsection{Evaluation}
\label{sec:setup:eval}

\paragraph{Benchmarks.}
We evaluate on two long-horizon tool-use benchmarks:
(i) \textbf{AppWorld} \citep{appworld}, a stateful environment spanning 9 apps and 457 APIs evaluated via programmatic unit tests. We report on its two held-out splits, \awtn (\texttt{test\_normal}, 168 tasks) and \awtc (\texttt{test\_challenge}, 417 tasks featuring unseen apps and composition patterns).
(ii) \textbf{$\tau$-bench Banking} \citep{taubench}, evaluating multi-turn customer service in the banking domain (\texttt{all-tools} setting, using \texttt{GPT-5.4} as the user simulator and judge).
Crucially, \textbf{we train exclusively on AppWorld}; $\tau$-bench is a zero-shot transfer benchmark. Ground-truth success signals are used only for evaluation, never during training.

\paragraph{Metrics.}
On AppWorld, success is measured by \textbf{Task Goal Completion (TGC)} (unit test pass rate) and \textbf{Scenario Goal Completion (SGC)} (full-scenario success). On $\tau$-bench, success is measured by the state-matching \textbf{Success Rate (SR)}.
To evaluate both discovery and consistency across $n{=}3$ runs, we report $\text{pass@}k$ (\textit{discovery}: $\ge 1$ success in $k$ trials) and $\text{pass}^k$ (\textit{consistency}: success in all $k$ trials) as defined by \citet{taubench}; at $k{=}1$ both equal mean success. We write $p^k$ for $\text{pass}^k$ and report $p^1$, $p^2$, and $p^3$ over the full task split (unsolved tasks count as failures).
We also report two efficiency metrics: average \textbf{agent turns} per episode and evaluation \textbf{cost} in USD.\footnote{Calculated via token usage at reference prices per 1M in/out tokens: \$0.289/\$0.9 for \texttt{Qwen3.6-27B}; \$0.66/\$0.8 for \texttt{Qwen2.5-32B-Instruct}; \$2.5/\$15 for \texttt{GPT-5.4}.}

\input{tables/table_main}

\paragraph{Evaluation protocol.}
We train \texttt{Qwen3.6-27B} for 100 steps (20 epochs) and \texttt{Qwen2.5-32B-Instruct} for 75 (15 epochs). For each setting, we report the mean over the final three epoch checkpoints, running three independent evaluation runs per checkpoint using AppWorld's official harness and Simplified ReAct Code Agent scaffold. We avoid selecting checkpoints by held-out task performance to prevent leaking implicit supervision \citep{salt}. Episodes are capped at 50 turns on both AppWorld splits and 100 on $\tau$-bench.

%% file: tables/table_main.tex
\begin{table*}[!t]
\centering
\small
\setlength{\tabcolsep}{4pt}
\renewcommand{\arraystretch}{1.15}
\resizebox{\textwidth}{!}{%
\begin{tabular}{l!{\vrule}ccc@{\hspace{8pt}}ccc!{\vrule}ccc@{\hspace{8pt}}ccc!{\vrule}ccc!{\vrule}ccc}
\toprule
\multirow{3}{*}{\textbf{Setting}}
& \multicolumn{6}{c!{\vrule}}{\textbf{AppWorld}$_\text{TN}$}
& \multicolumn{6}{c!{\vrule}}{\textbf{AppWorld}$_\text{TC}$}
& \multicolumn{3}{c!{\vrule}}{$\boldsymbol{\tau}_\text{B}$}
& \multicolumn{3}{c}{} \\
\cmidrule(lr){2-7} \cmidrule(lr){8-13} \cmidrule(lr){14-16}
& \multicolumn{3}{c}{\textbf{TGC}} & \multicolumn{3}{c!{\vrule}}{\textbf{SGC}}
& \multicolumn{3}{c}{\textbf{TGC}} & \multicolumn{3}{c!{\vrule}}{\textbf{SGC}}
& \multicolumn{3}{c!{\vrule}}{\textbf{SR}}
& \multicolumn{3}{c}{\multirow{-2}{*}{\textbf{Average}}} \\
\cmidrule(lr){2-4} \cmidrule(lr){5-7} \cmidrule(lr){8-10} \cmidrule(lr){11-13} \cmidrule(lr){14-16} \cmidrule(lr){17-19}
& $p^1$ & $p^2$ & $p^3$ & $p^1$ & $p^2$ & \multicolumn{1}{c!{\vrule}}{$p^3$}
& $p^1$ & $p^2$ & $p^3$ & $p^1$ & $p^2$ & \multicolumn{1}{c!{\vrule}}{$p^3$}
& $p^1$ & $p^2$ & \multicolumn{1}{c!{\vrule}}{$p^3$}
& $p^1$ & $p^2$ & $p^3$ \\
\midrule
\rowcolor{blockhl}
\multicolumn{19}{@{}l}{\texttt{Qwen2.5-32B-Instruct}} \\
Base                 & 35.7 & 27.4 & 23.2 & 17.3 & 11.3 & 8.9 & 17.3 & 10.6 & 7.4 & 5.8 & 2.9 & 2.2 & 3.4 & 1.7 & 1.0 & 15.9 & 10.8 & 8.5 \\
\rowcolor{verifhl}
SALT (Outcome-aware)                & 66.2 & - & - & 47.9 & - & - & 36.8 & - & - & 20.9 & - & - & - & - & - & - & - & - \\
\rowcolor{ourshl}
DRACO (Ours)         & 62.9 & 52.6 & 46.0 & 42.3 & 32.7 & 26.8 & 34.0 & 24.8 & 20.2 & 19.2 & 13.2 & 10.3 & 4.0 & 1.9 & 1.5 & 32.5 & 25.0 & 21.0 \\
\midrule
\rowcolor{blockhl}
\multicolumn{19}{@{}l}{\texttt{Qwen3.6-27B}} \\
Base                 & 69.4 & 55.2 & 47.6 & 41.1 & 28.0 & 21.4 & 49.7 & 38.3 & 31.7 & 30.2 & 18.7 & 12.9 & 15.8 & 9.6 & 7.2 & 41.2 & 30.0 & 24.2 \\
\rowcolor{verifhl}
Outcome reward & 80.0 & 70.0 & 63.3 & 59.3 & 45.0 & 38.1 & 59.9 & 47.2 & 40.4 & 38.3 & 27.7 & 21.3 & 17.6 & 11.0 & 9.3 & 51.0 & 40.2 & 34.5 \\
DRACO w/o Dyn.\ \& Cred. & 81.1 & 71.4 & 64.7 & 59.9 & 45.4 & 37.5 & 60.5 & \textbf{50.4} & \textbf{45.6} & \textbf{42.4} & \textbf{33.3} & \textbf{27.3} & 19.4 & 11.1 & 8.9 & 52.7 & 42.3 & 36.8 \\
DRACO w/o Dyn.          & 81.9 & 72.0 & 65.5 & 60.7 & 45.6 & 39.3 & 59.4 & 48.1 & 41.9 & 39.2 & 28.4 & 21.8 & 18.7 & 10.9 & 7.9 & 52.0 & 41.0 & 35.3 \\
DRACO w/o Cred.         & 82.1 & 73.1 & 66.3 & 64.9 & 52.2 & 44.6 & 59.3 & 48.4 & 42.5 & 40.6 & 31.3 & 26.4 & 19.7 & 11.9 & 9.3 & 53.3 & 43.4 & 37.8 \\
\rowcolor{ourshl}
DRACO (Ours)              & \textbf{85.3} & \textbf{78.1} & \textbf{72.8} & \textbf{70.6} & \textbf{59.5} & \textbf{51.8} & \textbf{61.5} & 49.7 & 43.9 & 40.7 & 32.3 & \textbf{27.3} & 20.4 & 12.0 & 8.6 & \textbf{55.7} & \textbf{46.3} & \textbf{40.9} \\
\rowcolor{ourshl}
DRACO (Ours - self-judge) & 81.1 & 72.1 & 65.5 & 62.7 & 48.2 & 38.1 & 61.0 & 45.6 & 36.9 & 34.7 & 21.7 & 16.1 & \textbf{21.1} & \textbf{14.7} & \textbf{11.7} & 52.1 & 40.5 & 33.7 \\
\bottomrule
\end{tabular}}
\caption{\textbf{Task success and consistency} (\%) on \awtn (168 tasks), \awtc (417 tasks),
$\tau_\text{B}$ (97 tasks). $p^k$ denotes $\text{pass}^k$ over $n{=}3$ runs. Average is a row-level summary only. Trained variants are compared to their block's Base. \colorbox{verifhl}{Outcome-aware} comparisons and \colorbox{ourshl}{our methods} are highlighted. DRACO ablations remove dynamic rubrics or step credit.
-- = not evaluated on that split, or not reported by the cited work.}
\label{tab:main}
\end{table*}

%% file: sections/06-results.tex
\subsection{Overall Results}
\label{sec:results}

We report mean over a run's last three checkpoints. Additional results are in Appendix~\ref{sec:appendix}.

\paragraph{Gains over baselines.} DRACO consistently outperforms the untrained policy models \texttt{Qwen3.6-27B} and \texttt{Qwen2.5-32B-Instruct}. On \texttt{Qwen3.6-27B}, it improves \awtn TGC/SGC from $69.4/41.1$ to $85.3/70.6$ ($+15.9/+29.5$) and \awtc TGC from $49.7$ to $61.5$ (as shown in Table \ref{tab:main}). It achieves the best results on all reported \awtn metrics and the highest average across benchmarks at every consistency level, with larger gains at higher consistency levels (e.g., \awtn $p^3$ TGC: $72.8$ vs.\ $47.6$). With \texttt{Qwen2.5-32B-Instruct}, DRACO raises \awtn TGC/SGC from $35.7/17.3$ to $62.9/42.3$ ($+27.2/+25.0$), the larger gain of the two base policy models, and closes most of the distance to the outcome-aware baseline SALT~\cite{salt} for step-wise credit assignment ($66.2/47.9$). Note that SALT uses the ground-truth reward signal and their method needed special processing to be applied to AppWorld due to its continuous textual state and action space. DRACO also transfers to $\tau$-bench without training, raising SR from $15.8$ to $20.4$, despite using no verifier, gold answers, or reference trajectories.

\paragraph{Consistency and discovery.} Table~\ref{tab:main} reports $p^1$--$p^3$, while Figure~\ref{fig:passk} compares consistency with $\text{pass@}k$. On \awtn, DRACO improves both discovery and consistency, with substantially larger gains in consistency: TGC $p^3$ increases by $25.2$ points, whereas $\text{pass@}3$ rises by only $3.9$. The untrained policy model could often solve tasks at least once; DRACO makes those successes reliable across repeated attempts. Retention ($p^3/p^1$) shows the same pattern, with DRACO achieving the highest retention on both TGC and SGC.

\begin{figure}[!b]
\centering
\includegraphics[width=\columnwidth]{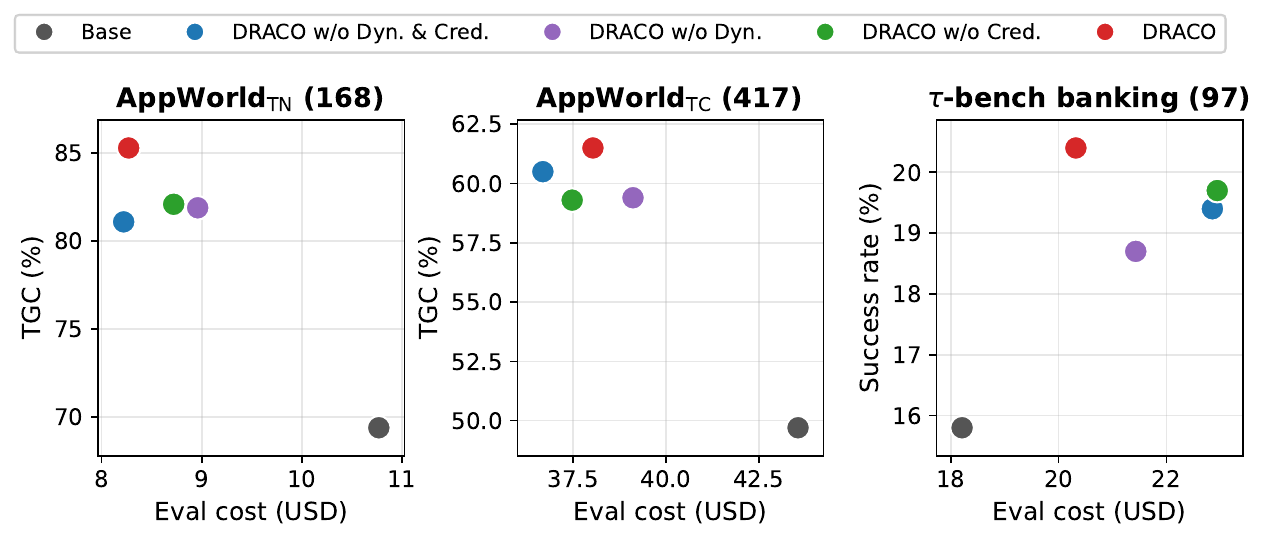}
\caption{\textbf{Performance vs.\ evaluation cost}, one panel per benchmark; up and to the left
is better. AppWorld panels plot TGC, $\tau$-bench banking plots success rate; SGC follows
the same ordering (Table~\ref{tab:main}). Cost is the US-dollar total for one pass over the
full split at the rates of Section~\ref{sec:setup:eval}; the $\tau$-bench axis also includes
the \texttt{GPT-5.4} user simulator and judge.
Points are \texttt{Qwen3.6-27B} $p^1$ values from Table~\ref{tab:main}.}
\label{fig:pareto}
\end{figure}

\subsection{Ablations}
\label{sec:ablations}

We ablate along three axes: whether the reward sees task outcomes at all, which of our
two components is responsible for the outcome-blind regime's gains, and whether
the judge has to be a frontier model.

\paragraph{Outcome reward.}
The outcome-reward setting is trained on AppWorld's unit tests instead of a judge using vanilla GRPO keeping the 
same base model and hyperparameters.
Per Table \ref{tab:main}, DRACO outperforms it by $+5.3$ TGC and $+11.3$ SGC on
\awtn and $+1.6$ TGC and $+2.4$ SGC on \awtc. The margin widens with
\emph{consistency}: on \awtn it reaches $+9.5$ TGC and $+13.7$ SGC at $p^3$, and
\awtc widens the same way ($+3.5$ TGC, $+6.0$ SGC). 
Training on random rewards has been reported to improve performance in some
settings \citep{spuriousrewards}. Doing so here reaches $74.0$ TGC and $50.0$ SGC on
\awtn, well below DRACO's $85.3$ and $70.6$.

\paragraph{Rubric type and step credit.}
Per Table \ref{tab:main}, DRACO leads every ablated setting on the $p^1$ average ($55.7$ against $52.0$--$53.3$) and
by a wider margin at $p^3$ ($40.9$ against $35.3$--$37.8$). The four-way comparison locates
that gain in the \emph{combination} of the two components. On \awtn, adding
step credit to per-trajectory rubrics is worth $+3.2$ TGC and $+5.7$ SGC, and the two
components together are worth $+4.2$ TGC and $+10.7$ SGC over the fixed rubric alone,
growing to $+8.1$ TGC and $+14.3$ SGC at $p^3$. Either
component on its own contributes far less ($+0.8$ and $+1.0$ TGC), which is what the
mechanism predicts: step credit needs criteria specific enough to implicate particular
steps, and a rubric written for the whole task distribution gives the attributor little to
attribute. \awtc sharpens the interaction into a sign change. There step
credit on a fixed rubric costs $3.7$ TGC at $p^3$, while on per-trajectory rubrics it adds
$+1.4$ TGC, so the same intervention pays off precisely once the rubric is specific to the
episode. The gains hold across task difficulty (Table~\ref{tab:difficulty-std}): on
\awtn, step credit on per-trajectory rubrics improves TGC at every level and
is largest on \emph{medium} tasks ($+7.2$ TGC, $+14.6$ SGC), and it lifts \emph{easy}
scenario completion by $+8.2$ SGC. All four settings also transfer to $\tau$-bench, gaining
between $+2.9$ and $+4.6$ SR over the untrained policy model.

\begin{figure*}[t]
\centering
\includegraphics[width=\textwidth]{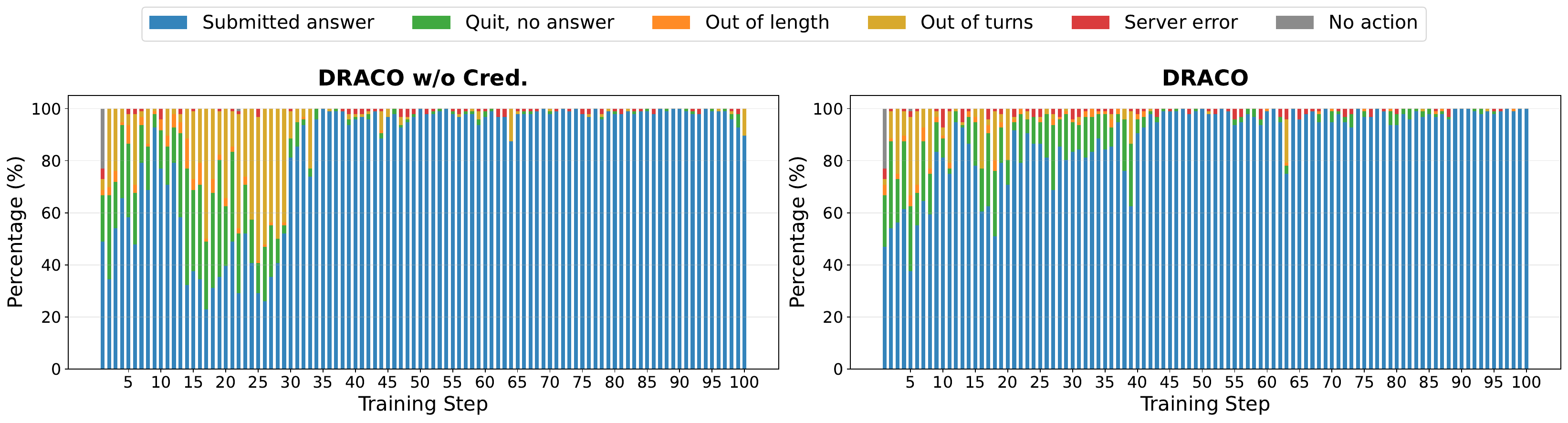}
\caption{\textbf{Rollout termination modes across training steps for DRACO with and without step credit.} Bars show the percentage of rollouts per termination mode (normal completion/give-up vs.\ length, turn, or server errors), independent of task correctness. We interpret broad trends rather than exact error distributions.}
\label{fig:terminations}
\end{figure*}
\begin{figure}[!tb]
\centering
\includegraphics[width=\columnwidth]{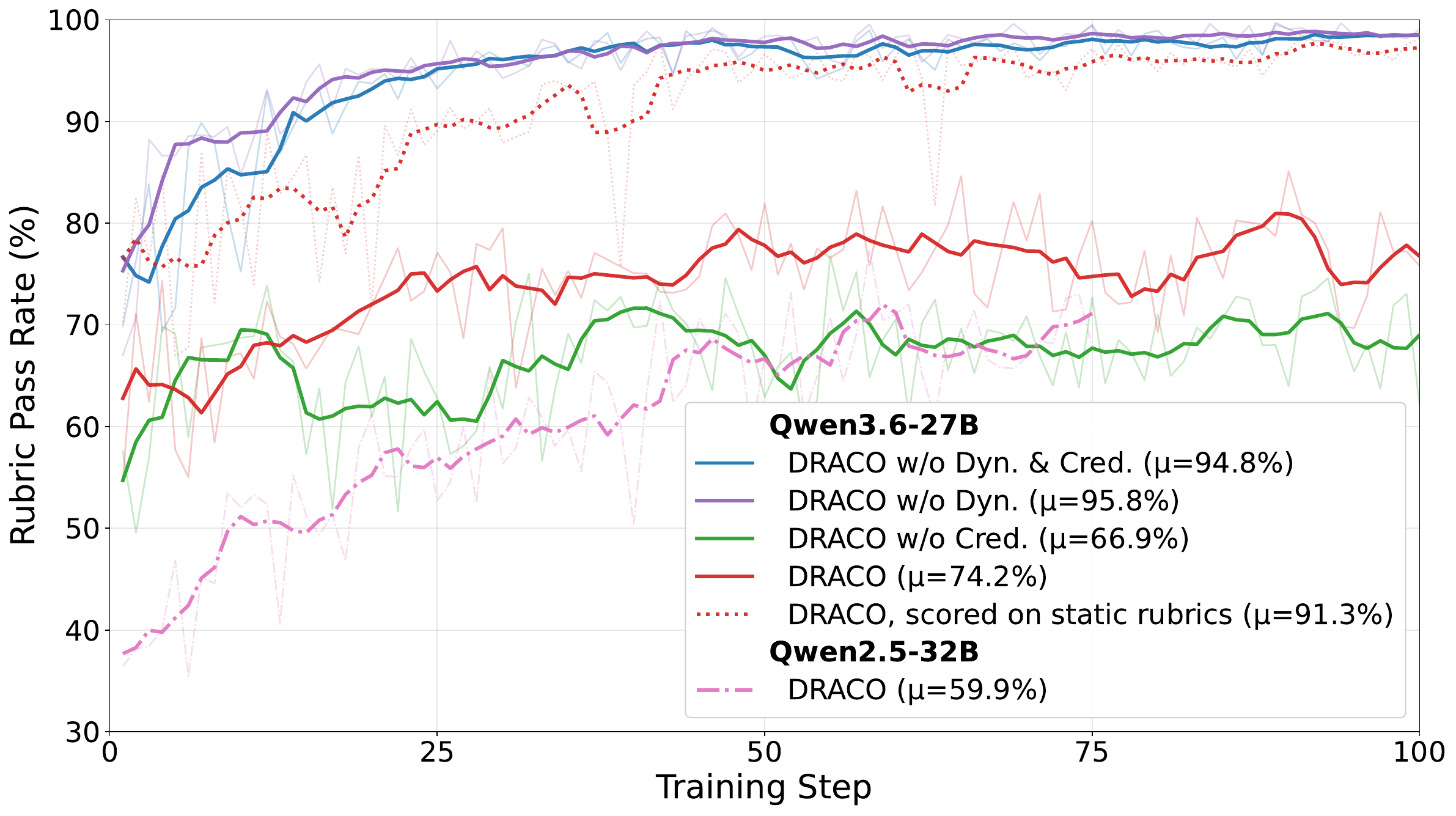}
\caption{\textbf{Overall rubric pass rate across training steps.} Each setting is evaluated on its training rubrics (dynamic rubrics vary per step). The dotted line shows DRACO re-scored on the static held-out rubric set. Figure~\ref{fig:perrubric} in Appendix~\ref{sec:appendix} reports per-criterion pass rates.}
\label{fig:passrate}
\end{figure}

\paragraph{Self-judge.}
Every setting above scores with a frozen frontier judge, which is the single largest
cost in the pipeline. We replace it with the policy model judging its own rollouts, with
thinking enabled in every judge phase; scoring is repeated $k{=}3$ times per trajectory,
and a criterion counts as passed only if all three calls pass it, so a single lenient
call cannot result in a pass. Appendix~\ref{sec:appendix:selfjudge-variants} compares this
judge with a variant that disables thinking and scores with a single call. This cuts the
judge cost for $100$ training steps from \$1607 to \$316, a $5.1\times$ saving, and the
saving holds across all five phases that issue judge calls
(Figure~\ref{fig:judgecost} in the Appendix). On \awtn it exceeds the outcome-aware reference, $81.1/62.7$
TGC/SGC against its $80.0/59.3$ and $65.5/38.1$ against $63.3/38.1$ at $p^3$, and on
$\tau$-bench it achieves the best SR of any setting ($21.1$): a policy
model grading its own rollouts reaches verifier-trained performance without a verifier.
Replaying every scoring call through the policy checkpoint, with a single call and no thinking,
shows the verdicts largely agree with
the frontier judge's ($89.4$\% of $60{,}689$ criterion-level verdicts, against $72.0$\% for a
judge that passes everything), and that where they disagree the self-judge is
\emph{lenient} rather than noisy: it passes $30.4$\% of the criteria the frontier judge failed
while failing only $1.3$\% of the ones it passed
(Appendix~\ref{sec:appendix:analysis:selfjudge}).

\input{tables/table_efficiency}

\subsection{Analysis}
\label{sec:analysis}
\label{sec:analysis:sub}

We turn from scores to what the gains cost and how training produced
them: evaluation cost and episode length (Figure~\ref{fig:pareto},
Table~\ref{tab:efficiency}), rollout termination
(Figure~\ref{fig:terminations}), and reward progression during training
(Figure~\ref{fig:passrate}, and Figure~\ref{fig:traintgc} in
Appendix~\ref{sec:appendix}). Appendix~\ref{sec:appendix:analysis} has further analysis.

\paragraph{Evaluation cost and episode length.}
All four rubric settings on both AppWorld splits
in Figure~\ref{fig:pareto}, have higher accuracy gains with lower evaluation cost
rather than \emph{Base}. On \awtn, DRACO reaches
$85.3$ TGC for \$8.27 against $69.4$ at \$10.77 untrained, and on \awtc
$61.5$ for \$38.03 against $49.7$ at \$43.56. The untrained policy model is thus the most
expensive split to evaluate despite scoring lowest.
Appendix~\ref{sec:appendix:frontier} compares DRACO's performance and evaluation cost
with those of frontier models on \awtn.
Table~\ref{tab:efficiency} shows where that saving comes from. On \texttt{Qwen3.6-27B}, every rubric setting shortens the episode on both AppWorld splits, DRACO
by the most ($18.7$ to $14.7$ turns on \awtn, $22.9$ to $20.7$
on \awtc), so it is solving more tasks in fewer agent turns rather than
buying accuracy with a longer rollout. On $\tau$-bench the picture reverses: turns rise
slightly for every setting and the gains cost more, which is what we would expect on a
benchmark held out of training, where the untrained policy model's shorter episodes reflect
giving up earlier rather than finishing sooner.

\paragraph{How rollouts end.}
Figure~\ref{fig:terminations} shows how each rollout terminated over the course of training.
Both dynamic settings converge on the same behaviour: the agent almost always submits an
answer by the end, and the modes where the episode simply ran out (response length
exceeded, turn budget exhausted, an environment server failure, or no action taken at
all) fade to negligible. These errors dominate early training and adding
credit clears them sooner: dynamic alone spends first $30$ steps oscillating, at times
dropping below $30$\% submitted, whereas DRACO settles within roughly
the first $40$. 

\paragraph{Static rubrics saturate; dynamic ones do not.}
Figure~\ref{fig:passrate} shows the reward the policy model actually optimizes. Against static
rubrics it saturates early, reaching the mid-$90$s within about $25$ steps and
staying there, so the two static settings are indistinguishable for most of training
($94.8$\% and $95.8$\% mean) and the reward stops discriminating long before training ends.
The dynamic settings sit lower and keep moving ($66.9$\% and $74.2$\%). This is by design
rather than a sign of worse rollouts: the generator is prompted for criteria at least one
member of the group is likely to fail, and discriminative dropout then discards the rest
(Section~\ref{sec:method:rubric}). Step credit helps here too, with DRACO better than
DRACO w/o Cred.\ for nearly all of training.
Furthermore, re-scoring DRACO's rollouts against the
fixed static rubrics they never trained on recovers $91.3$\%, close to the static
runs' own curve, so per-trajectory rubrics subsume the static criteria. Figure \ref{fig:passrate} shows on the weaker base policy model the pass rate climbs steeply from below $40$\% to around $75$\%, the
largest change of any run in the figure; thus, the signal is not confined to a strong
base policy model.

%% file: tables/table_efficiency.tex
\begin{table}[!b]
\centering
\small
\setlength{\tabcolsep}{3pt}
\renewcommand{\arraystretch}{1.15}
\resizebox{\columnwidth}{!}{%
\begin{tabular}{@{}l!{\vrule}c!{\vrule}c!{\vrule}c!{\vrule}c@{}}
\toprule
\textbf{Setting}
& \textbf{AppWorld}$_\text{TN}$
& \textbf{AppWorld}$_\text{TC}$
& $\boldsymbol{\tau}_\text{B}$
& \textbf{Avg.} \\
\midrule
\rowcolor{blockhl}
\multicolumn{5}{@{}l}{\texttt{Qwen2.5-32B-Instruct}} \\
Base                 & 16.4 & 23.8 & 21.5 & 20.6 \\
\rowcolor{ourshl}
DRACO (Ours)         & 19.6 & 21.0 & 27.1 & 22.6 \\
\midrule
\rowcolor{blockhl}
\multicolumn{5}{@{}l}{\texttt{Qwen3.6-27B}} \\
Base                 & 18.7 & 22.9 & \textbf{27.7} & 23.1 \\
\rowcolor{verifhl}
Outcome reward & 17.4 & 23.7 & 29.6 & 23.6 \\
DRACO w/o Dyn.\ \& Cred. & 15.9 & 21.2 & 30.1 & 22.4 \\
DRACO w/o Dyn.          & 15.6 & 21.6 & 29.3 & 22.2 \\
DRACO w/o Cred.         & 16.4 & 22.2 & 30.1 & 22.9 \\
\rowcolor{ourshl}
DRACO (Ours)              & \textbf{14.7} & \textbf{20.7} & 27.9 & \textbf{21.1} \\
\rowcolor{ourshl}
DRACO (Ours - self-judge) & 17.4 & 24.6 & 28.9 & 23.6 \\
\bottomrule
\end{tabular}}
\caption{\textbf{Average agent turns}, lower is better. -- = not evaluated on that split.}
\label{tab:efficiency}
\end{table}

%% file: sections/08-conclusion.tex
\section{Conclusion}
\label{sec:conclusion}
We study reinforcement learning for long-horizon tool-using agents in the outcome-blind setting, where ground-truth success signal is not available and rewards must come entirely from process criteria. We introduce two components: task-specific rubrics generated from policy model rollouts and merged at the group level, and a step-credit assignment rule that redistributes trajectory advantage using the judge's per-criterion attributions while preserving its total magnitude and sign. 
Our method improves AppWorld's TGC by 15.9 points on \awtn, transfers zero-shot to $\tau-$bench, and matches or exceeds a verifier-trained baseline with the same budget. Ablations show that step-level credit assignment on top of per-trajectory rubrics is key;  the main challenge is making evaluation criteria specific enough to support meaningful attribution. Since the judge defines the objective rather than estimating it, understanding the required judge quality is a natural direction for future work.

%% file: sections/09-limitations.tex

A limitation inherent to using rubrics as rewards on unverifiable tasks is that 
there is no independent signal against which to check whether the criteria describe it
faithfully. Validating the criteria would require human raters, which we leave to future work. We use a single frozen judge for most of our experiments, and do not report a
chance-corrected measure of its per-criterion agreement with human
annotators~\citep{reliabilitywithoutvalidity}. A judge that is internally consistent can still
be systematically wrong, and our experiments cannot separate the two.

The same gap appears one level down, in the attribution. We validate DRACO by end-task
performance, which does not establish that credit lands on the right steps: a
redistribution that credits the wrong steps can still improve a policy model, and one that
credits the right steps can fail to.

Another characteristic of DRACO is that discriminative dropout makes the surviving criterion set a function of the sampled group:
a criterion is kept only if some member failed it, so the same prompt can be scored
against different criteria at different points in training. We do not measure the
resulting variance; our repeated runs are inference-time repeats of fixed checkpoints,
so training-time variability is not characterized.


%% file: sections/10-ethics.tex
Both benchmarks, AppWorld \citep{appworld} and $\tau$-bench \citep{taubench}, are
used under their released licenses and are fully simulated: every app, account and
record is a synthetic fixture, so no real user data or live service is touched and
no human subjects were involved. Because the reward is a set of natural-language
criteria, the target of optimization is explicit and auditable, which we regard as
a safety property; the corresponding risk is that any bias the judge carries is
inherited by the policy with no verifier to catch it. 
Agents that use tools more competently are dual-use: the same capability that completes a
booking or a refund can execute an unwanted transaction more effectively but our
contribution is to the training signal rather than to autonomy or tool access, and
all evaluation is sandboxed.

%% file: sections/11-appendix.tex
\section{Additional Results}
\label{sec:appendix}

\begin{figure*}[t]
\centering
\includegraphics[width=\textwidth]{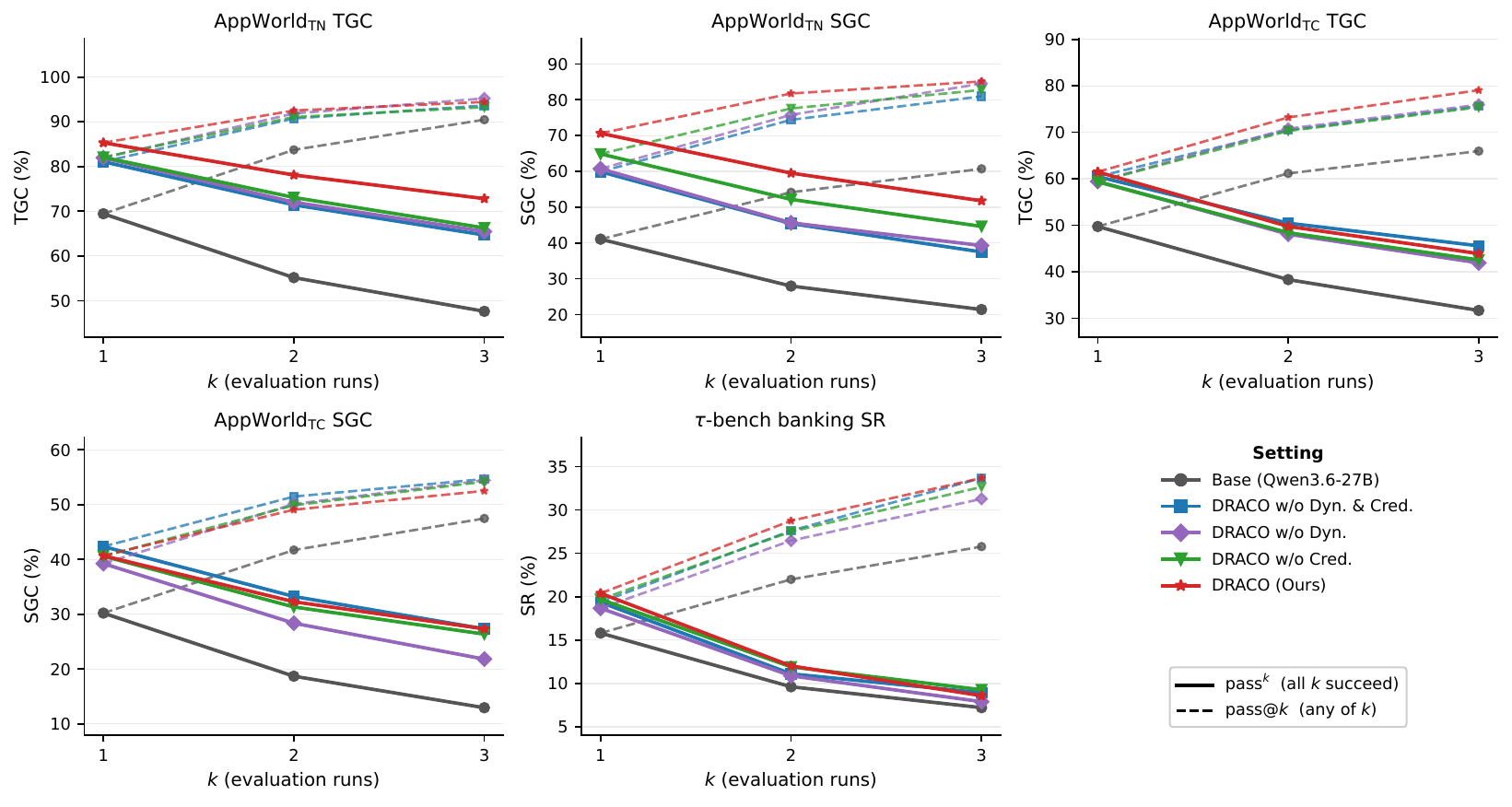}
\caption{\textbf{$\text{pass}^k$ vs.\ $\text{pass@}k$} over $n{=}3$ evaluation runs, in \%.
Solid: $\text{pass}^k$, all $k$ runs succeed. Dashed: $\text{pass@}k$, at least one
succeeds. The two coincide at $k{=}1$ by construction. Values and protocol follow
Table~\ref{tab:main}. All panels are \texttt{Qwen3.6-27B}; axes do not start at zero.}
\label{fig:passk}
\end{figure*}

\begin{figure}[t]
\centering
\includegraphics[width=\columnwidth]{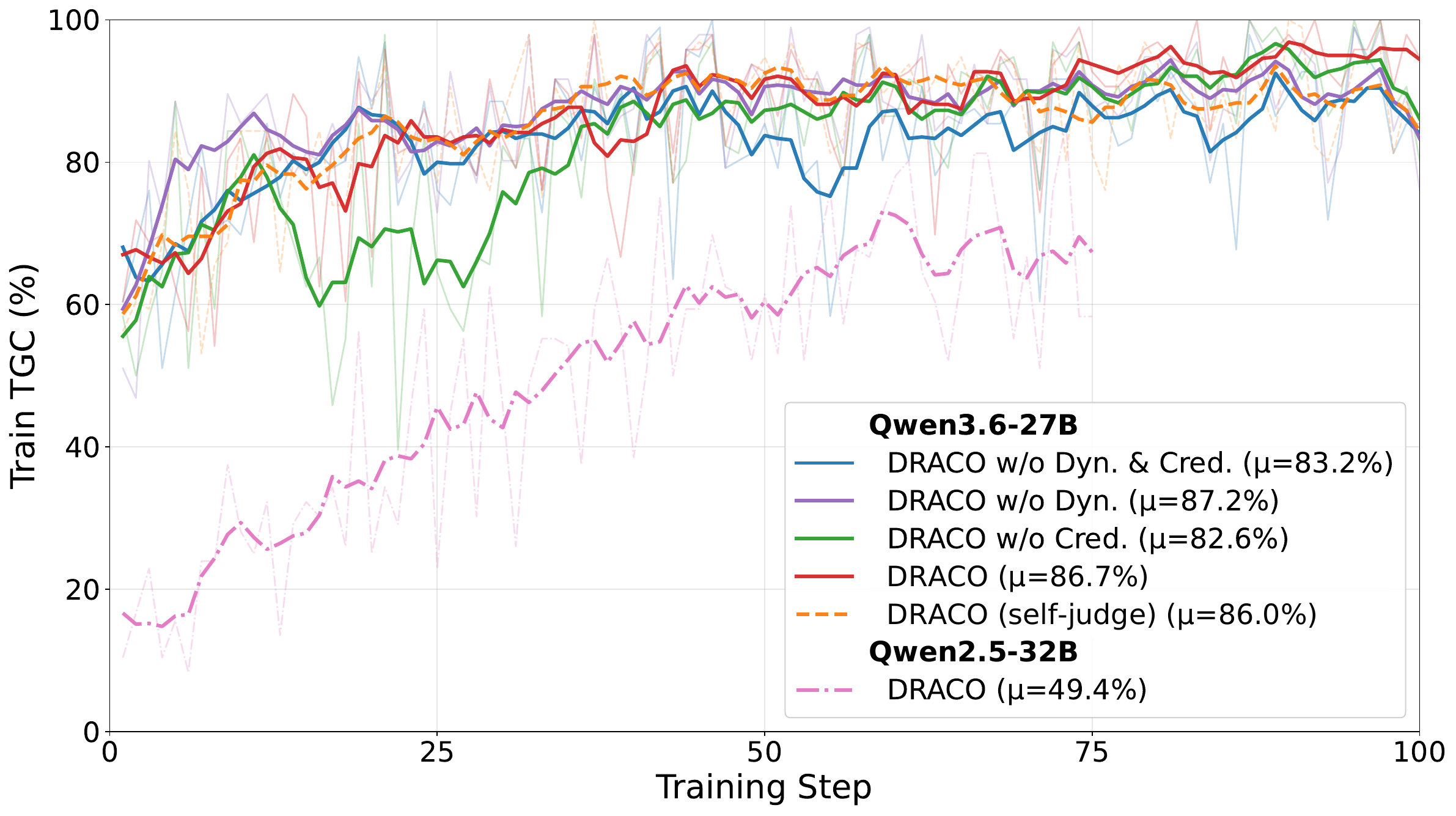}
\caption{\textbf{Train TGC vs.\ training step} This is TGC on the \emph{training} tasks, for
post-hoc analysis, and it is not part of any reward.}
\label{fig:traintgc}
\end{figure}

\begin{figure*}[t]
\centering
\begin{subfigure}[t]{0.78\textwidth}
\centering
\includegraphics[width=\textwidth]{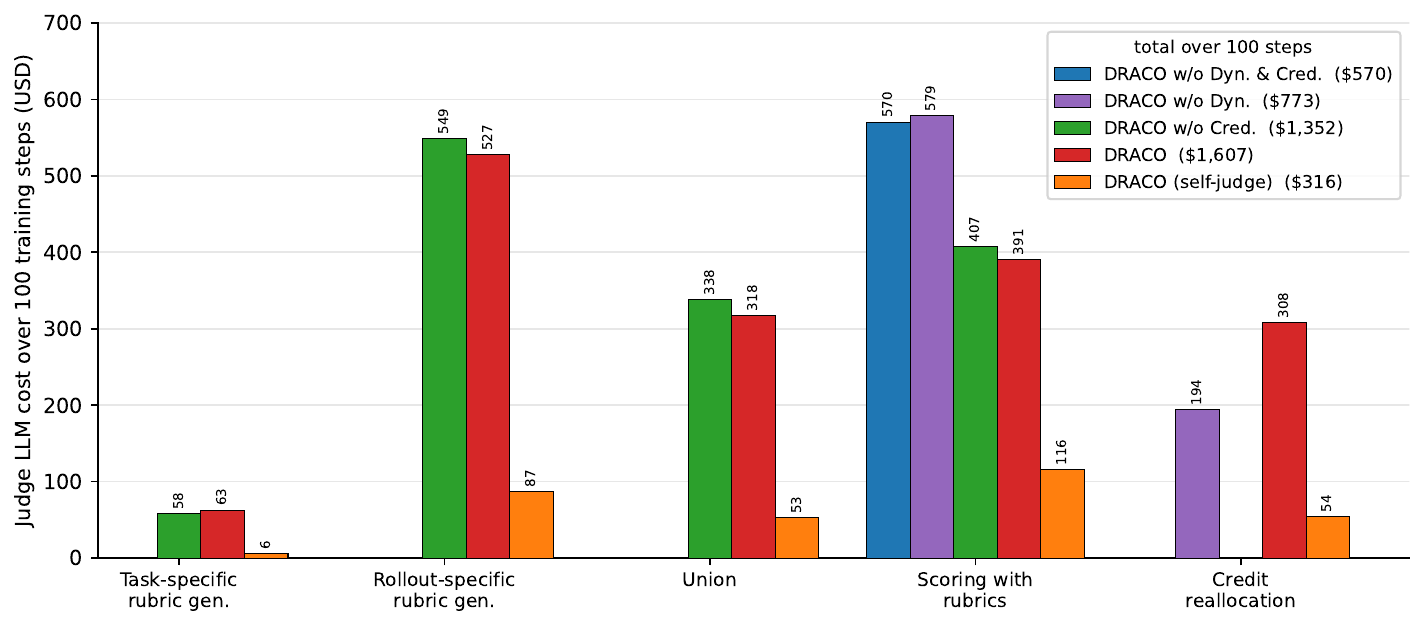}
\caption{Cost by pipeline phase.}
\label{fig:judgecost:phase}
\end{subfigure}

\vspace{1.2em}
\begin{subfigure}[t]{0.46\textwidth}
\centering
\includegraphics[width=\textwidth]{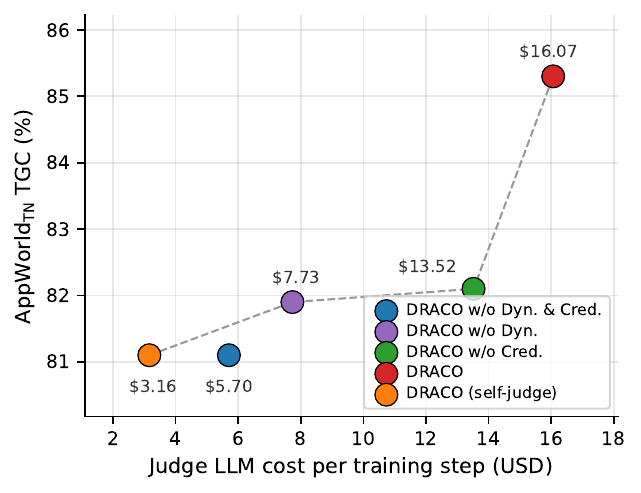}
\caption{Cost against \awtn TGC.}
\label{fig:judgecost:pareto}
\end{subfigure}
\caption{\textbf{Training-time judge cost}, in USD, for the four outcome-blind settings and
the self-judge variant, which scores with the policy model itself, with thinking enabled
and $k{=}3$ scoring calls per trajectory (Section~\ref{sec:ablations}). All settings use
\texttt{Qwen3.6-27B} as the policy model and \texttt{GPT-5.4} as the judge, except the
self-judge. (a) Cost over $100$ training steps, split across the five phases that issue judge
calls: task-specific and rollout-specific rubric generation, union, scoring, and credit
reallocation. A setting that does not run a phase draws no bar. (b) The same cost per
training step against \awtn TGC ($p^1$, Table~\ref{tab:main}); up and to the left is better.
Costs use the token rates of Section~\ref{sec:setup:eval}.}
\label{fig:judgecost}
\end{figure*}

\begin{figure*}[t]
\centering
\includegraphics[width=\textwidth]{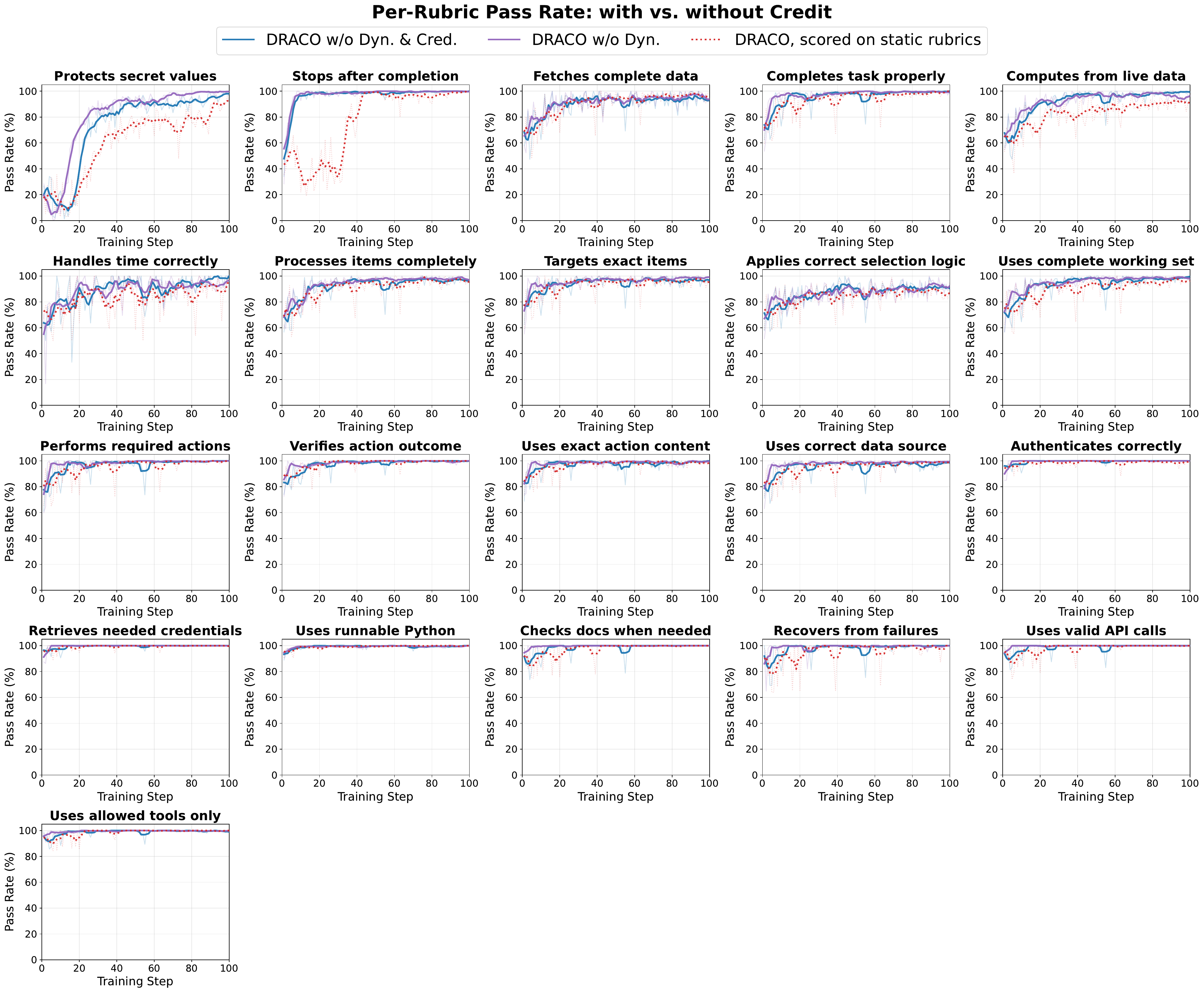}
\caption{\textbf{Per-rubric pass rate vs.\ training step}, static rubrics with and without step
credit. One panel per static criterion; faint lines
are raw per-step values, bold lines smoothed. The dotted series re-scores DRACO
against the same fixed set, which it never trained on.}
\label{fig:perrubric}
\end{figure*}

This section holds breakdowns that the main text summarizes but does not have room to
show. Tables~\ref{tab:main-std} and~\ref{tab:efficiency-std} restate Tables~\ref{tab:main}
and~\ref{tab:efficiency} with the standard deviation across the three checkpoints each
reported mean averages. Table~\ref{tab:difficulty-std} splits task success by AppWorld's own
difficulty label on the same basis, and is where the by-difficulty reading in
Section~\ref{sec:ablations} comes from.
Figure~\ref{fig:passk} plots $\text{pass}^k$ against $\text{pass@}k$ for every setting,
separating consistency from discovery (Section~\ref{sec:results}).
Figure~\ref{fig:traintgc} tracks TGC on the training tasks over the course of training, an
optimization signal rather than a held-out metric.
Figure~\ref{fig:judgecost} breaks the training-time judge cost down by pipeline phase for a
frontier judge and for the self-judge variant.
Figure~\ref{fig:perrubric} opens up the aggregate static pass rate of
Figure~\ref{fig:passrate} one criterion at a time: most of the 21 static criteria are
already near ceiling within $20$ steps, and the aggregate curve's early rise comes from a
few initially-hard criteria, chief among them \emph{Protects secret values}, which starts
below $20$\% and is where step credit shows its clearest per-criterion advantage.

%

\input{tables/table_main_std}
\input{tables/table_efficiency_std}
\input{tables/table_difficulty_std}

\begin{figure*}[!t]
\centering
\includegraphics[width=\textwidth]{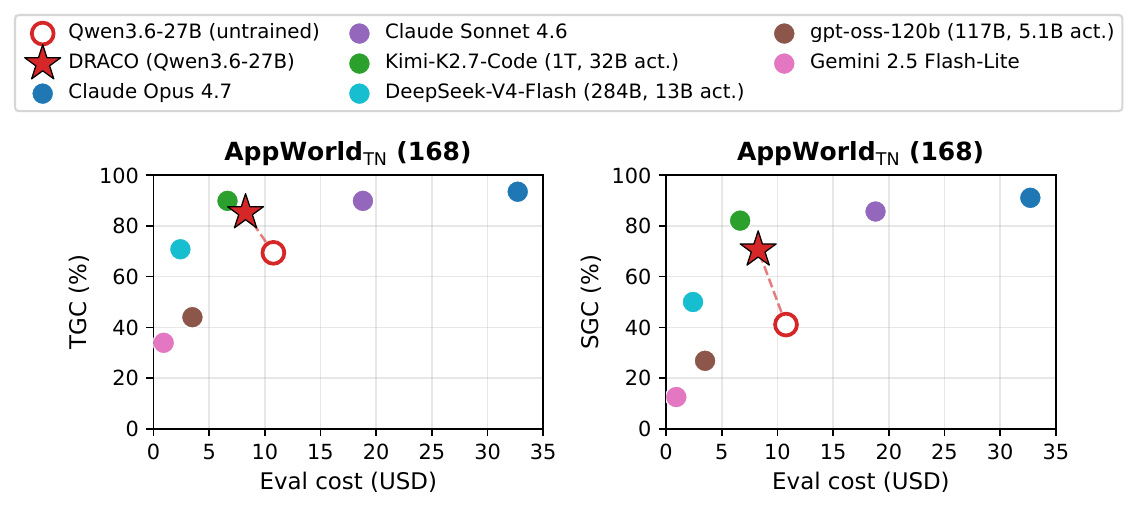}
\caption{\textbf{Frontier models vs.\ DRACO on \awtn.} TGC (left) and SGC (right) against
the cost of one pass over the $168$ tasks; up and to the left is better. Circles: frontier
models, one run each. Star: DRACO on
\texttt{Qwen3.6-27B} with the frontier judge. Hollow circle: the same policy untrained.
Brackets give parameter counts (total, active) for open-weight models; closed models publish none.
Values are in Table~\ref{tab:frontier}.}
\label{fig:frontier}
\end{figure*}

\begin{figure*}[t]
\centering
\includegraphics[width=\textwidth]{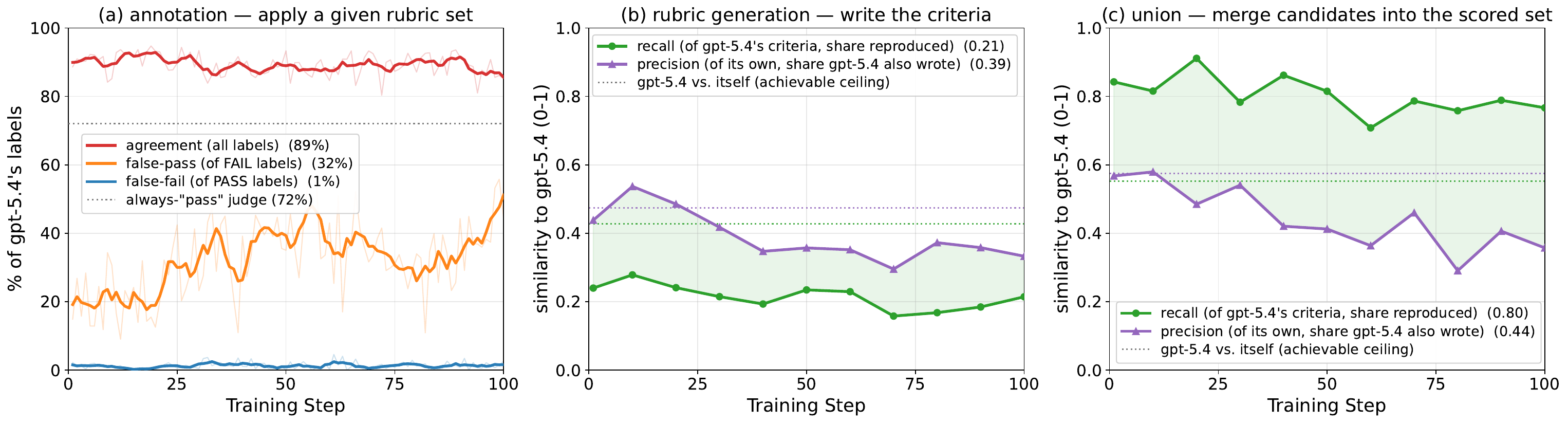}
\caption{\textbf{The three jobs a judge does, self-model vs.\ \texttt{GPT-5.4}}, measured by
replaying the run's judge calls through the policy checkpoint on the archived prompts, with a
single call and no thinking (the ablated self-judge of Table~\ref{tab:selfjudge-variants}).
\textbf{(a)~Scoring a rubric it is given.} How often the self-model returns the same verdict as
\texttt{GPT-5.4}, over all $60{,}689$ verdicts, and the two ways it can differ. Each series has its
own denominator, named in the legend. Faint lines are per-step values, bold lines smoothed.
\textbf{(b)~Writing the criteria} and \textbf{(c)~merging them} into the set a group is scored on.
Here the two models write their own criteria, so we compare the sets themselves: a third model pairs
up criteria that mean the same thing, and we then ask how much the two sets overlap. \emph{Recall} is
how much of \texttt{GPT-5.4}'s set the self-model also wrote; \emph{precision} is how much of the
self-model's set \texttt{GPT-5.4} also wrote. Legends give means, and the shading is simply the gap
between the two curves. Perfect overlap is not the target, because two judges writing criteria for
the same task will not pick the same checks: the dotted lines show \texttt{GPT-5.4} compared against
\emph{itself} on two rollouts of one task, and are the reference to read the curves against.
The self-model scores a given rubric well but too leniently, reproduces only about a fifth of
the criteria \texttt{GPT-5.4} writes, and at the merge keeps most of \texttt{GPT-5.4}'s
criteria while adding roughly twice as many of its own.}
\label{fig:selfjudge}
\end{figure*}

\subsection{Self-judge variants}
\label{sec:appendix:selfjudge-variants}

The self-judge of Section~\ref{sec:ablations} enables thinking in every judge phase and
repeats scoring $k{=}3$ times per trajectory, counting a criterion as passed only if all
three calls pass it. Table~\ref{tab:selfjudge-variants} compares it against an earlier
variant with neither change: no thinking, and a single scoring call. The unanimity rule
targets the self-judge's dominant error, lenient false passes
(Appendix~\ref{sec:appendix:analysis:selfjudge}).

\input{tables/table_selfjudge}

\subsection{Comparison with frontier models}
\label{sec:appendix:frontier}

Figure~\ref{fig:frontier} and Table~\ref{tab:frontier} compare DRACO with six frontier and
open-weight models on \awtn, plotting TGC and SGC against evaluation cost as in
Figure~\ref{fig:pareto}. The frontier models are run once each, with no training on
AppWorld.\footnote{Legend sizes come from the model cards of the open-weight models;
\texttt{Claude} and \texttt{Gemini} publish none.} DRACO is the \texttt{Qwen3.6-27B} frontier-judge setting of
Table~\ref{tab:main}, shown next to its untrained policy.

Our training brings a $27$B model close to models one to two orders of magnitude larger,
after $100$ steps on outcome-blind rewards alone. It lifts the policy from $69.4/41.1$ TGC/SGC
to $85.3/70.6$, and lowers its evaluation cost from \$10.77 to \$8.27. That places DRACO
within $4.6$ TGC of \texttt{claude-sonnet-4-6} and $8.2$ of \texttt{claude-opus-4-7}, at
$44$\% and $25$\% of their cost, and ahead of \texttt{DeepSeek-V4-Flash} ($284$B) and
\texttt{gpt-oss-120b} by $14.5$ and $41.3$ TGC. Only \texttt{Kimi-K2.7-Code}, a $1$T
model, is both cheaper and stronger. The largest gain is on SGC, where every task in a
scenario must pass: training raises it by $+29.5$ points, from below
\texttt{DeepSeek-V4-Flash} to within $11.5$ of \texttt{Kimi-K2.7-Code}.

\input{tables/table_frontier}

\section{Additional Analysis}
\label{sec:appendix:analysis}

\subsection{Self-judge vs.\ frontier judge}
\label{sec:appendix:analysis:selfjudge}

The self-judge variant replaces \texttt{GPT-5.4} with the policy model. To measure what that costs
we replay the run's judge calls rather than compare two trained runs, which would confound the
judge with the policy it shaped. At each step $N$ we re-send that step's calls to the
step-$(N{-}1)$ checkpoint, using the archived prompt verbatim, so the only thing that changes is
which model answers. \texttt{GPT-5.4}'s output is the reference: every number below measures
agreement with it, not correctness. This replay predates the $k{=}3$-with-thinking judge used for
the self-judge row of Table~\ref{tab:main}; every self-judge number in this section is a single
call with no thinking, so it is a lower bound on what that stronger judge agrees on.

The judge does three jobs, and the self-model is good at one of them
(Figure~\ref{fig:selfjudge}). It applies a rubric nearly as well as \texttt{GPT-5.4}. It writes
different criteria. And when it merges criteria into the set a group is scored on, it keeps the
right ones but cannot discard the rest.

\subsubsection{Annotation: applying a given rubric}

This is the easy job and the self-model does it well. Over $100$ steps, $9{,}590$ rollouts and
$60{,}689$ verdicts, it matches \texttt{GPT-5.4} on $89.4$\% of them
(Figure~\ref{fig:selfjudge}a). A judge that answers ``pass'' every time would match $72.0$\%, so
the agreement is not just an artifact of pass-skewed labels. Verdicts the self-model failed to
produce count as disagreements: it left $1.3$\% of them unanswered, and \texttt{GPT-5.4} left none.

The errors run one way. The self-model passes $30.4$\% of the criteria \texttt{GPT-5.4} failed
($4{,}817$ of $15{,}840$), but fails only $1.3$\% of those it passed ($560$ of $43{,}706$) --- a
lenient error is about $24$ times more likely than a strict one. That is the better failure mode to
have, because passing a criterion wrongly narrows the reward spread within a rollout group rather
than reordering the group, and GRPO learns from that spread. But leniency grows over training: the
false-pass rate rises from $22.1$\% over the first quarter to $34.9$\% over the last
($p{=}2\mathrm{e}{-}6$), while the false-fail rate stays flat and below $5$\% at every step
($p{=}0.34$).

Agreement itself barely moves ($90.8$\% to $89.0$\% between the first and last quarter). We
attribute that small drop to the criteria rather than to the judge: over the same span the criteria
per rollout fall from $7.1$ to $5.8$, and the share \texttt{GPT-5.4} itself fails falls from
$31.5$\% to $23.8$\%, so later steps ask for finer calls over fewer failures. Both are properties
of the generator and of the frontier judge's own labels. (The figure's legend averages over steps;
the totals here pool every verdict, so the two differ by up to two points.)

\subsubsection{Generation: writing the criteria}

Applying a rubric is not the same as writing one, so we replayed the generation calls too. Here
there is no reference label to match, so we measure overlap instead: a third model (Claude Sonnet)
matches the two arms' criteria one to one, which gives recall --- how many of \texttt{GPT-5.4}'s
criteria the self-model also wrote --- and precision --- how many of its own \texttt{GPT-5.4} also
wrote. This covers $1{,}024$ rollouts over $11$ steps.

An overlap of $1.0$ is not the target, because two authors given one task pick different checks. We
therefore score \texttt{GPT-5.4} against \emph{itself} on the same metric, comparing its criteria
for two different rollouts of the same task, and get recall $0.43$ and precision $0.48$. Those are
the dotted lines in Figure~\ref{fig:selfjudge}b, and they are what ``as similar as \texttt{GPT-5.4}
is to itself'' looks like.

The self-model stays below them. Recall averages $0.21$, half the ceiling, and sits between $0.16$
and $0.28$ at every step. Precision averages $0.39$ and declines from $0.44$ to $0.33$
($p{=}0.006$). Precision is above recall at all $11$ steps, and the self-model writes fewer
criteria than \texttt{GPT-5.4} ($3.8$ against $5.1$ per rollout): what it writes usually resembles
something \texttt{GPT-5.4} wrote, but it misses most of what \texttt{GPT-5.4} wrote.

\subsubsection{Union: merging candidates into the scored set}

A group's final rubric set is merged down from the candidates the previous two stages
proposed, and this is where the substitution breaks. We score the merges the same way, over
$175$ groups; the ceiling here is recall $0.55$ and precision $0.57$
(Figure~\ref{fig:selfjudge}c).

Recall averages $0.80$ and is above that ceiling at all $11$ steps, so the self-model keeps four in
five of the checks \texttt{GPT-5.4} kept --- it is not throwing away the right criteria. Precision
averages $0.44$ and falls from $0.57$ to $0.36$ ($p{=}0.002$), below recall at every step. The
counts say why: the self-model keeps $15.3$ criteria per group where \texttt{GPT-5.4} keeps $6.3$.
It finds the right criteria and then keeps far too much else alongside them.

That surplus costs reward signal. Scoring both arms' merged sets on the same six rollouts with one
fixed scorer, the share of criteria that at least one rollout fails --- the criteria that survive
our discriminativeness filter and so produce a within-group gradient --- is lower for the
self-model at all $11$ steps ($31.3$\% against $46.5$\%). The extra criteria are ones every rollout
passes, so they lengthen the rubric without adding signal.

Two caveats bound these readings. The replay is pinned to the archived prompts, so the self-model's
generation extends \texttt{GPT-5.4}'s proposals and its merge consumes \texttt{GPT-5.4}'s candidate
pool; a fully self-judged run would feed weaker candidates into a weaker merge, and that compounding
is not measured here. And the two ceilings rest on $24$ and $14$ scored pairs respectively, with the
union ceiling built from the same task at different steps because union makes only one call per task
per step, so both are scale references rather than exact bounds.

\subsection{What the dynamic rubrics contain}
\label{sec:appendix:analysis:naprobe}

Section~\ref{sec:analysis:sub} shows that DRACO's rollouts recover $91.3$\% on the fixed static set
they never trained on, so the per-trajectory criteria subsume the static ones. This section asks the
complementary question: what do they add? We tally every criterion the judge scored over all $96$
rollouts of DRACO's $99$ completed steps ($6{,}245$ distinct criteria, $60{,}119$ judgments), and
label each one by how many of the $90$ training tasks it was ever written for.

\paragraph{Almost every criterion is a sub-goal of one task.}
$84.4$\% of the distinct criteria are written for a single task, and they carry $57.2$\% of all
judgments; only $18$ criteria are shared across ten or more tasks. A static set has no such register
by construction: with one set covering all $90$ tasks, every criterion must be phrased generally
enough to apply everywhere. So the two regimes differ less in how many questions they ask
than in \emph{what kind}: for the Spotify task \emph{``Follow all artists of all indie-genre songs
in any of my playlists''}, DRACO writes \emph{Indie songs were identified correctly},
\emph{Deduplicates artists before follow}, and \emph{Only own playlists used}; for
\emph{``Accept all pending Venmo payment requests from my coworkers and friends''} it writes
\emph{Used contacts for eligibility}, \emph{Checks funding before approval}, and
\emph{Approved all and only eligible requests}. None is expressible in a rubric shared with the other
task, and each names a specific way that instruction can be misread --- following artists from
someone else's playlist, approving a request from a non-coworker.

\paragraph{The recurring criteria are the static ones, rediscovered.}
The $18$ task-general criteria fall into four families: not exposing secrets (seven phrasings),
handling pagination (five), stopping after completion (three), and recovering from an API or
execution error (three). All four are already in the static set (\emph{Protects secret values},
\emph{Fetches complete data}, \emph{Stops after completion}, \emph{Recovers from failures};
Table~\ref{tab:staticrubrics}), which is the subsumption result seen from the generator's side: given
only the instruction and a rollout, it re-derives the same cross-cutting concerns a human-curated set
was built around, and does so on the tasks where they bite rather than on all of them.

\paragraph{The task-specific criteria are the ones that discriminate.}
$26.7$\% of DRACO's applicable verdicts are failures, against $5.5$\% for the static set, so the
per-trajectory criteria are far harder to satisfy --- which is what the reward needs, since a
criterion the whole group passes contributes nothing to the advantage after
group normalization (Eq.~\eqref{eq:grpo-adv}). The same asymmetry appears in what the judge can
score at all: a static criterion draws \emph{not applicable} on $10.7$\% of judgments and a
per-trajectory one on $1.9$\%, because a set shared across $90$ tasks must include criteria that fit
only some of them (\emph{Handles time correctly} is inapplicable on $67.6$\% of the rollouts it was
scored against, and six criteria account for $95.6$\% of all such verdicts).

\subsection{Why step credit needs dynamic rubrics}
\label{sec:appendix:analysis:inert}

Section~\ref{sec:ablations} finds that step credit is worth $+3.2$ TGC on per-trajectory rubrics but
only $+0.8$ on the static set. The credit rule itself explains why. It reallocates a trajectory's
advantage only when the rubric's verdicts are \emph{mixed} --- some criteria pass and others fail on
the same trajectory. If they are unanimous, every step takes the same weight and the update is
exactly baseline GRPO
(Appendix~\ref{sec:appendix:credit:allpass}); the same is true when discriminative dropout leaves no
criterion at all. Credit is then \emph{inert}: present in the pipeline, has no effect on the gradient.

Because the condition is exact, we can count it. Over each arm's $100$ steps we classify all $96$
rollouts per step from the logged verdicts, and average the inert share within five windows of
training. On the static set it climbs from $23.6$\% of rollouts over steps $1$--$10$ to $55.4$\%,
$82.0$\%, $88.9$\% and $91.9$\% by steps $76$--$100$. DRACO starts at a similar $29.7$\% and stays
near it: $44.7$\%, $44.5$\%, $52.9$\% and $56.5$\%. Averaged over training, credit is inert on
$76.4$\% of static rollouts against $48.1$\% of DRACO's, and the spread of step qualities within a
trajectory --- how unevenly credit divides the advantage when it does fire --- is correspondingly
larger for DRACO ($0.215$ against $0.144$).

So the static arm does not attribute badly; for most of training it does not attribute at all. The
rise mirrors the saturation of Figure~\ref{fig:passrate}: once the policy passes almost every fixed
criterion, unanimous rollouts are the common case, and by the last quarter of training the credit
component is switched off on nine rollouts in ten. Per-trajectory rubrics keep producing criteria
some group member fails, which is what leaves credit something to divide. The pattern is not specific
to one policy model: the \texttt{Qwen2.5-32B-Instruct} run under DRACO moves only from $15.7$\% to
$30.9$\% inert over the same windows.

This measures when credit is inert, which follows from the rule, not whether an active reallocation
lands on the steps truly responsible --- the attribution-quality question we flag in the Limitations
section.

\section{Rubric Construction}
\label{sec:appendix:rubrics}

The static rubric set was authored once before training and then frozen. We sampled 8
rollouts for each of the 90 training tasks from the \emph{untrained} base policy model (720
trajectories at temperature $1.0$), had \texttt{GPT-5.4} propose criteria per task, reduced them to
a single set, and pruned criteria that were rarely applicable or rarely discriminative.
The dynamic settings instead generate criteria per task at every step during training
(Section~\ref{sec:method:rubric}).

Every generation call, static and dynamic alike, instructs the judge to keep the criteria
MECE (Section~\ref{sec:method:rubric}), and the
pruning above enforces the same property on the static set after the fact: a criterion
that is rarely applicable is not covering a dimension the tasks actually exercise, and one
that is rarely discriminative is usually restating a criterion already in the set.
Table~\ref{tab:staticrubrics} lists the resulting 21 criteria, ordered by their failure
rate over those 720 rollouts. Nothing was pruned at this stage:
the pruning happened while reducing the per-task proposals to the single set. Applicability
survived unevenly --- $13$ of the $21$ apply to at least $99\%$ of those rollouts, while $8$ apply
to fewer, down to $29.9\%$ for \emph{Handles time correctly}, and those $8$ are what
Appendix~\ref{sec:appendix:analysis:naprobe} measures the cost of at training time. The
failure rates span two orders of magnitude, from $81.8\%$ on \emph{Protects secret values}
down to $2.8\%$ on \emph{Uses allowed tools only}, which is what discriminative dropout
(Section~\ref{sec:method:rubric}) exists to handle at training time: the criteria near the
bottom of the table pass for almost every group and are dropped from the reward.

\input{tables/table_static_rubrics}

\section{Training Hyperparameters}
\label{sec:appendix:hparams}

Table~\ref{tab:hparams} gives the full training configuration
(Section~\ref{sec:setup:training}). It is shared by all four settings, which differ only
in the reward (Section~\ref{sec:setup:settings}); the credit rule's own knobs are in
Appendix~\ref{sec:appendix:credit:config}. Advantages are the within-group
standardized rewards of Eq.~\ref{eq:grpo-adv}, so the group mean is the only baseline
and no value network is trained; the 16 prompts and $G{=}6$ rollouts each give 96
rollouts per step, and every run trains for at most 20 epochs (100 optimizer steps)
with a checkpoint written at every step. Every run trains on 8 NVIDIA H100 GPUs.

Three choices are worth flagging because they are easy to assume otherwise. First, we
train LoRA adapters rather than the full model, so the reported gains come from a
rank-16 update on top of frozen base weights. Second, we run a single PPO epoch per
batch, so every update is on-policy and the importance ratio is $1$ at the time of the
step, which makes the clipping range inactive in practice. Third, the learning rate is
constant at $5\!\times\!10^{-5}$ with no warmup, so no schedule interacts with the
last-three-checkpoint reporting protocol (Section~\ref{sec:setup:eval}).

\begin{table}[h]
\centering
\small
\setlength{\tabcolsep}{4pt}
\begin{tabular}{@{}ll@{}}
\toprule
\textbf{Setting} & \textbf{Value} \\
\midrule
\multicolumn{2}{@{}l}{\textit{Objective}} \\
\quad Advantage estimator & GRPO \\
\quad Group-std normalization & yes (Eq.~\ref{eq:grpo-adv}) \\
\quad Value network & none; group mean \\
\quad KL penalty & loss term, $k_3$ \citep{schulmankl} \\
\quad KL coefficient & $0.01$ \\
\addlinespace
\multicolumn{2}{@{}l}{\textit{Batching}} \\
\quad Prompts per step & 16 \\
\quad Group size $G$ & 6 \\
\quad Rollouts per step & 96 \\
\quad Minibatch size & 8 \\
\quad PPO epochs per batch & 1 (on-policy) \\
\quad Training length & 20 epochs (100 steps) \\
\quad Checkpoint interval & every step \\
\addlinespace
\multicolumn{2}{@{}l}{\textit{Optimizer}} \\
\quad Optimizer & AdamW \\
\quad Learning rate & $5\!\times\!10^{-5}$ \\
\quad Schedule & constant, no warmup \\
\quad Weight decay & $0.01$ \\
\quad Gradient clipping & $1.0$ \\
\addlinespace
\multicolumn{2}{@{}l}{\textit{Policy loss}} \\
\quad Clip range (low / high) & $0.2$ / $0.2$ \\
\quad Dual-clip constant & $3.0$ \\
\quad Entropy bonus & $0$ \\
\quad Loss aggregation & token-mean \\
\addlinespace
\multicolumn{2}{@{}l}{\textit{Rollout sampling}} \\
\quad Temperature & $1.0$ \\
\quad Top-$p$ / top-$k$ & $1.0$ / unrestricted \\
\quad Sampling & stochastic \\
\quad Max assistant turns & 50 \\
\addlinespace
\multicolumn{2}{@{}l}{\textit{Sequences}} \\
\quad Max prompt length & 4096 \\
\quad Max response length & 24576 \\
\quad Truncation side & left \\
\quad Overlong prompts & filtered \\
\addlinespace
\multicolumn{2}{@{}l}{\textit{Trainable parameters}} \\
\quad Adaptation & LoRA \citep{lora} \\
\quad Rank / $\alpha$ & 16 / 32 \\
\quad Target modules & all linear \\
\quad Base models & \texttt{Qwen3.6-27B}, \\
                 & \texttt{Qwen2.5-32B-Instruct} \\
                 & (both frozen) \\
\addlinespace
\multicolumn{2}{@{}l}{\textit{Hardware}} \\
\quad GPUs per run & 8$\times$NVIDIA H100 \\
\bottomrule
\end{tabular}
\caption{\textbf{Training hyperparameters}, shared by all four settings and by both
base policies.}
\label{tab:hparams}
\end{table}

\paragraph{Serving configuration for \texttt{Qwen2.5-32B-Instruct}.}
We serve \texttt{Qwen2.5-32B-Instruct} at its native 32{,}768-token context, using its own
chat template and a tool-call parser matching the JSON format it emits. On $\tau$-bench we
extend the window to 131{,}072 tokens with YaRN rope scaling \citep{yarn}. Thinking mode is
off, since its chat template defines no reasoning channel.

\paragraph{$\tau$-bench retrieval.}
Both base policies retrieve with \texttt{Qwen} embeddings in the \texttt{all-tools} setting.

\section{Credit Reallocation: Details}
\label{sec:appendix:credit}

This appendix restates the preliminaries and gives the full derivation of the
credit-reallocation rule (Section~\ref{sec:method:credit}), the significance of
each term, a worked example, a traced real rollout, and the configuration used in
our runs.

\subsection{Preliminaries: GRPO and the token-level advantage}
\label{sec:appendix:credit:prelim}

We train with Group Relative Policy Optimization (GRPO). For a task $x$ the policy model
samples a group of $G$ trajectories $\{\tau_i\}_{i=1}^{G}$, and each trajectory
$\tau_i$ receives a scalar reward $R_i$. GRPO standardizes rewards within the
group to form the trajectory advantage
\begin{equation}
  A_i = \frac{R_i - \mu_g}{\sigma_g},
  \label{eq:app-grpo-adv}
\end{equation}
where $\mu_g$ and $\sigma_g$ are the mean and standard deviation of the group's
rewards. Writing the tokens of $\tau_i$ as $y_{i,1}, \dots, y_{i,N_i}$, GRPO takes
a policy-gradient step in which every token is pushed by the same advantage,
\begin{equation}
  \nabla_\theta \mathcal{J}
  = \mathbb{E}\!\left[\, \sum_{t=1}^{N_i} A_i \,
    \nabla_\theta \log \pi_\theta(y_{i,t} \mid y_{i,<t}, x) \right].
  \label{eq:app-grpo-pg}
\end{equation}
So the advantage acts as a per-token scalar multiplier on that token's
log-probability gradient: a positive advantage makes the token more likely in its
context, a negative advantage less likely, and a zero advantage leaves it
untouched. The defining limitation is that a single $A_i$ multiplies all $N_i$
tokens, so every decision in a long trajectory receives identical credit. Credit
reallocation keeps the GRPO reward and group standardization intact and replaces
only the uniform multiplier: every token in step $j$ receives a per-step advantage
$a_j$ (Eq.~\eqref{eq:app-credit}) in place of $A_i$ in Eq.~\eqref{eq:app-grpo-pg}.

\subsection{Setup and notation}
\label{sec:appendix:credit:setup}

A trajectory $i$ is a multi-turn agent rollout, decomposed into steps
$j = 1, \dots, M$, where one step is approximately one agent turn (one emitted
code block). Each response token belongs to exactly one step, or to a
\emph{gap} (turn glue, tool-result echoes) that belongs to no step. Two quantities
enter the assignment:
\begin{itemize}
  \item $A_i$, the trajectory-level GRPO advantage of
  Eq.~\eqref{eq:app-grpo-adv}, whose sign says whether the rollout is reinforced
  or suppressed and whose magnitude says how strongly.
  \item $Q_j \in [0,1]$, the quality of step $j$, derived from the rubric grader
  (Eq.~\eqref{eq:app-stepq}).
\end{itemize}
Credit is assigned at the step level: the rubric grades steps, not tokens, so
there is no token-resolution signal to reallocate on. The scheme replaces the flat
$A_i$ with a per-step advantage $a_j$ that depends on the step's quality, and every
token inside step $j$ receives the same $a_j$.

\subsection{Formulation}
\label{sec:appendix:credit:rule}

Let $p_i$ and $f_i$ count the passed and failed criteria over the whole trajectory
$\tau_i$ (surviving dropout and applicable), let $p_j$ and $f_j$ count those citing step
$j$, and let $n_j$ be the number of tokens in step $j$, with $N = \sum_k n_k$. The
assignment, from rubric reward through to the step advantage that replaces $A_i$
(restating Sections~\ref{sec:method:rubric} and~\ref{sec:method:credit}), is
\begin{align}
  R_i &= \frac{p_i - f_i}{p_i + f_i},
    \label{eq:app-reward} \\[4pt]
  A_i &= \frac{R_i - \mu_g}{\sigma_g},
    \label{eq:app-adv} \\[4pt]
  Q_j &= \frac{p_j}{p_j + f_j},
    \label{eq:app-stepq} \\[4pt]
  w_j &=
    \begin{cases}
      Q_j, & A_i \ge 0,\\
      1 - Q_j, & A_i < 0,
    \end{cases}
    \label{eq:app-stepw} \\[4pt]
  a_j &= A_i \cdot \frac{N\, w_j}{n_j \sum_k w_k},
    \label{eq:app-credit}
\end{align}
where $\mu_g$ and $\sigma_g$ are the group's reward mean and standard deviation
(Eq.~\eqref{eq:app-adv} repeats Eq.~\eqref{eq:app-grpo-adv} so the chain reads without
a jump), and $R_i = 0$ when no criterion applies. A step cited by no
criterion takes $Q_j = \bar{Q}$, the mean $Q$ over cited steps. Gap tokens receive
advantage $0$ and are excluded from $N$, so conservation holds over exactly the
tokens credit is distributed across. Every token in step $j$ takes the advantage
$a_j$ in place of $A_i$ in Eq.~\eqref{eq:app-grpo-pg}.

\subsection{Significance of each term}
\label{sec:appendix:credit:terms}

\paragraph{$A_i$ (trajectory advantage): sets the total push.}
The method is a redistribution of $A_i$, not a replacement of it. The sign of
$A_i$ says whether this rollout should be reinforced (winner, $A_i \ge 0$) or
suppressed (loser, $A_i < 0$) relative to its group; its magnitude says how
strongly. Credit only decides where within the trajectory that push lands.

\paragraph{$Q_j$ (step quality): the per-step signal.}
$Q_j$ (Eq.~\eqref{eq:app-stepq}) is the pass fraction of the rubric checks that cite
step $j$, where $p_j$ and $f_j$ are the number of passed and failed citing checks.
A step cited by no rubric check inherits $\bar{Q}$, the mean $Q$ over cited steps.
Evidence per step is thin: measured on the credit run, the number of rubrics
citing a single step has median $1$ and mean $1.95$, and only $0.6\%$ of steps are
cited by ten or more rubrics. So the majority of cited steps take $Q_j \in \{0,1\}$
from a single verdict, and the rule is deliberately literal about the judge's
attribution rather than hedging it toward a neutral value.

\paragraph{$w_j$ (step weight): converts quality into direction-correct push.}
The winner/loser split in Eq.~\eqref{eq:app-stepw} is what makes credit sign-correct.
On a winning trajectory we reinforce the steps that were actually good, so weight
rises with $Q_j$; on a losing trajectory we suppress the steps that were actually
bad, so weight rises with $1 - Q_j$. A step whose citing checks unanimously
disagree with the trajectory's direction takes $w_j = 0$: on a winner, a step every
citing criterion failed contributes nothing to the update, and on a loser, a step
every citing criterion passed is not suppressed. Both are the intended reading of
the attribution.

\paragraph{$n_j$ and the $1/n_j$ factor: length handling.}
The quantity the rule holds length-independent is each step's \emph{total}
contribution, not its per-token push. Step $j$ contributes
\begin{equation}
  n_j\, a_j = A_i N \frac{w_j}{\sum_k w_k},
  \label{eq:app-steptotal}
\end{equation}
which depends on $w_j$ and not on $n_j$: a step earns influence by being judged
good, not by being long. The $1/n_j$ factor spreads that fixed total over however
many tokens the step contains, so a verbose step pushes each of its tokens
proportionally less hard. The alternative, dropping $1/n_j$, would equalize the
per-token push instead and let a long step accumulate a proportionally larger
total, which would reward verbosity.

\paragraph{$N$ and conservation: same total push as baseline GRPO.}
The factor $N$ is chosen so the total advantage over the trajectory is preserved.
Summing Eq.~\eqref{eq:app-steptotal} over steps,
\begin{equation}
  \sum_j n_j\, a_j = A_i N \frac{\sum_j w_j}{\sum_k w_k} = A_i N.
  \label{eq:app-conserve}
\end{equation}
Baseline GRPO assigns $A_i$ to each of those same $N$ tokens, so its total push is
also $A_i N$: only the distribution across steps changes. Gap tokens sit outside
this accounting entirely, receiving advantage $0$ and being excluded from $N$, so
the conserved quantity is the baseline total over the credited tokens rather than
over the full response. Note that this conserves the scalar sum $\sum_j n_j a_j$;
it does not conserve the total gradient vector, since moving advantage between
steps whose tokens have different gradients is precisely the intended effect.

\paragraph{From advantage to update.}
At \texttt{ppo\_epochs} $= 1$, the policy-gradient update contributed by a token
in step $j$ is $\Delta\theta = \eta \, a_j \, \nabla_\theta \log \pi_\theta(\text{token})$,
so $a_j$ is a scalar multiplier on that token's log-probability gradient:
$a_j > 0$ makes the token more likely in its context, $a_j < 0$ less likely, and
$a_j = 0$ leaves it untouched.

\subsection{Worked example}
\label{sec:appendix:credit:worked}

Consider a trajectory with $A_i = 1$ and two steps of equal quality
($w_1 = w_2 = 1$), with $n_1 = 2$ and $n_2 = 4$ tokens, so $N = 6$ and
$\sum_k w_k = 2$. Then $a_1 = 6(1)/(2 \cdot 2) = 1.50$ and
$a_2 = 6(1)/(4 \cdot 2) = 0.75$, so $n_1 a_1 = n_2 a_2 = 3$ and
$\sum_j n_j a_j = 6 = A_i N$. Equal quality buys equal total influence: the two
steps split the trajectory's push evenly despite one being twice as long, which is
why the shorter step's tokens each carry more.

Now let step $2$ be higher quality, $w_1 = 0.5$ and $w_2 = 1.0$, so
$\sum_k w_k = 1.5$. Step 1 gets $a_1 = 6(0.5)/(2 \cdot 1.5) = 1.00$ and step 2 gets
$a_2 = 6(1.0)/(4 \cdot 1.5) = 1.00$, giving $n_1 a_1 = 2$, $n_2 a_2 = 4$, and again
$\sum_j n_j a_j = 6 = A_i N$. The good step now claims twice the total influence of
the weak one ($4$ against $2$), in the ratio of their weights; here that happens to
land both per-token advantages at $1.00$, because step 2 is both twice as good and
twice as long.

\subsection{Running example (real logged rollout)}
\label{sec:appendix:credit:running}

The following is a real rollout from the credit run (step 11), traced from the raw
rubric verdicts to the per-step advantage, so every $Q_j$ and the uncited
fallback are the actual logged values. Two quantities are illustrative and labelled
where used: the per-step token counts $n_j$ (the log records qualities and citations,
not token spans), and $A_i$, which we set to $1$ for readability. Every $a_j$ below
scales linearly in $A_i$, so the choice fixes the units of the table and nothing else.

The rollout has $M = 7$ steps and takes the winner branch, meaning $A_i \ge 0$: its
reward exceeded its group's mean. It is worth noting what that does \emph{not} mean.
Three of its five checks failed, so by Eq.~\eqref{eq:app-reward} its own reward
is $R_i = (2 - 3)/5 = -0.2$; it is a winner because the rest of its group scored
below $-0.2$, not because it did well in absolute terms and not because it passed the
task, which the reward never observes
(Appendix~\ref{sec:appendix:credit:config}). Five rubric checks fired, each citing one
or more steps:

\begin{center}
\small
\begin{tabular}{cll}
\toprule
\textbf{Check} & \textbf{Verdict} & \textbf{Cites steps} \\
\midrule
R1 & \textsc{fail} & 4, 6 \\
R2 & \textsc{fail} & 5, 6 \\
R3 & \textsc{fail} & 3, 4, 5 \\
R4 & \textsc{pass} & 5 \\
R5 & \textsc{pass} & 1, 2, 3, 4, 5, 6 \\
\bottomrule
\end{tabular}
\end{center}

\paragraph{Stage 1: tally passes/fails per step.}
For each step $j$, count the citing checks that passed ($p_j$) and failed
($f_j$). Step 7 is cited by nothing.

\begin{center}
\small
\begin{tabular}{cccl}
\toprule
\textbf{Step} & $p_j$ & $f_j$ & \textbf{Cited by} \\
\midrule
1 & 1 & 0 & R5 (\textsc{pass}) \\
2 & 1 & 0 & R5 (\textsc{pass}) \\
3 & 1 & 1 & R5 (\textsc{pass}); R3 (\textsc{fail}) \\
4 & 1 & 2 & R5 (\textsc{pass}); R1, R3 (\textsc{fail}) \\
5 & 2 & 2 & R4, R5 (\textsc{pass}); R2, R3 (\textsc{fail}) \\
6 & 1 & 2 & R5 (\textsc{pass}); R1, R2 (\textsc{fail}) \\
7 & 0 & 0 & \emph{(uncited)} \\
\bottomrule
\end{tabular}
\end{center}

\paragraph{Stage 2: quality $Q_j = p_j/(p_j + f_j)$.}
$Q_1 = Q_2 = 1/1 = 1.000$; $Q_3 = 1/2 = 0.500$; $Q_4 = Q_6 = 1/3 = 0.333$;
$Q_5 = 2/4 = 0.500$. Steps 1 and 2 are cited only by the one check that passed, so
they take the maximum; steps 4 and 6 are cited by one passing and two failing
checks and land lowest. Note how thin the evidence behind $Q_1 = Q_2 = 1$ is: their
only citation is R5, a broad check spanning six steps, so a single verdict decides
both. That is representative rather than cherry-picked, since the median step draws
exactly one citation (Appendix~\ref{sec:appendix:credit:terms}), but it is the reason
the uncited-step fallback averages over cited steps rather than trusting any one of them.

\paragraph{Stage 3: uncited step 7 inherits $\bar{Q}$.}
$Q_7 = \bar{Q} = \tfrac{1}{6}(1.0 + 1.0 + 0.5 + 0.333 + 0.5 + 0.333) = 0.611$.

\paragraph{Stage 4: winner weights.}
This is a winner, so the weight is the quality itself, $w_j = Q_j$, for every step
(including the uncited step 7, $w_7 = \bar{Q} = 0.611$; an uncited step is still a step, unlike a gap, which is outside $N$ entirely).

\paragraph{Stage 5: step advantage.}
The share of the trajectory's push each step claims is the ratio of weights and
needs no token counts: steps 1 and 2 (the strongest, $w = 1.000$) claim
$1.000/0.333 = 3\times$ the total influence of the weakest steps (4 and 6,
$w = 0.333$). This is the entire behavioral effect: the good early steps are
reinforced harder than the steps the rubric flagged as faulty. Token counts enter
only in spreading each step's share over its own tokens. Taking an illustrative
$n = [40, 40, 30, 50, 30, 60, 60]$, so $N = 310$, with $\sum_k w_k = 4.278$ and
$A_i = 1$:

\begin{center}
\small
\begin{tabular}{lccrr}
\toprule
\textbf{Step} & $n_j$ & $w_j$ & $a_j$ & step total $n_j a_j$ \\
\midrule
1 & 40 & 1.000 & 1.812 & 72.47 \\
2 & 40 & 1.000 & 1.812 & 72.47 \\
3 & 30 & 0.500 & 1.208 & 36.23 \\
4 & 50 & 0.333 & 0.483 & 24.16 \\
5 & 30 & 0.500 & 1.208 & 36.23 \\
6 & 60 & 0.333 & 0.403 & 24.16 \\
7 & 60 & 0.611 & 0.738 & 44.29 \\
\midrule
$\Sigma$ & 310 & & & 310.00 \\
\bottomrule
\end{tabular}
\end{center}

Baseline GRPO would put $a_j = 1.0$ on all seven steps ($\sum_j n_j a_j = 310$).
Credit keeps the same total, $\sum_j n_j a_j = A_i N = 310$, but rebalances it:
the strong steps take totals of $72.47$ against $24.16$ for the flagged ones, a
$3\times$ gap matching their weights, while the uncited step sits between them.
Changing the token counts changes the per-token advantages but never the step
totals (they depend on $w_j$ only) and never the conserved total. The $a_j$ column is
therefore not a ranking of the steps: a short step spreads its share over fewer tokens
and so shows a larger per-token value, which is why the step totals rather than $a_j$
are the quantity to read across rows.

\subsection{Unanimous rollouts stop discriminating between steps}
\label{sec:appendix:credit:allpass}

Credit discriminates between steps only when the rubric's verdicts are mixed
within a trajectory. If every citing verdict is a \textsc{pass}, then
$f_j = 0$ and $Q_j = 1$ for every cited step, the uncited steps inherit
$\bar{Q} = 1$, and all weights are equal. Eq.~\eqref{eq:app-credit} then reduces to
$a_j = A_i N / (n_j M)$: every step claims an equal share $A_i N / M$ of the
trajectory's total push, and the only remaining variation is the $1/n_j$ spreading
of that share over each step's own tokens. The citation \emph{count} carries no
signal here, only the pass/fail composition does. The mirror case behaves the same
way: on a loser whose every citing verdict is a \textsc{fail}, all $Q_j = 0$ so all
$w_j = 1 - Q_j = 1$, again uniform.

One degenerate case needs an explicit fallback. If a winner's citing verdicts are
all \textsc{fail}, every weight is $w_j = Q_j = 0$, the normalizer $\sum_k w_k$
vanishes, and Eq.~\eqref{eq:app-credit} is undefined. The implementation detects
the collapsed normalizer and falls back to the uniform $A_i$, so such a trajectory
trains as baseline GRPO rather than not at all. The same guard covers a loser whose
verdicts all pass.

\subsection{Configuration}
\label{sec:appendix:credit:config}

\begin{center}
\small
\begin{tabular}{lll}
\toprule
\textbf{Knob} & \textbf{Symbol} & \textbf{Value} \\
\midrule
Length handling & $1/n_j$ & step total is length-free \\
Gap tokens & n/a & advantage $0$, outside $N$ \\
Uncited-step quality & n/a & $\bar{Q}$ over cited \\
\bottomrule
\end{tabular}
\end{center}

Two degenerate cases: if no step is cited there is no attribution to reallocate on
and the implementation falls back to the uniform $A_i$, i.e.\ baseline GRPO. And because
the reward is outcome-blind, the winner/loser branch is keyed on the sign of the
GRPO advantage $A_i$ (its group-relative reward), not on task success, so a
rollout can be a ``winner'' for the weighting branch while still failing the task,
and vice versa.

\subsection{Properties}
\label{sec:appendix:credit:properties}

We state the invariants that justify the formulation, each with a short proof.
Throughout, step $j$ has $n_j > 0$ tokens, $N = \sum_j n_j$ over the steps, and
weight $w_j \ge 0$ with $\sum_k w_k > 0$ (the collapsed case falls back to baseline,
Appendix~\ref{sec:appendix:credit:allpass}). It is convenient to write step $j$'s
\emph{total} contribution as a share of the trajectory's total push,
\begin{equation}
  n_j\, a_j = A_i N \, \rho_j,
  \qquad \rho_j := \frac{w_j}{\sum_k w_k} \; (\ge 0),
  \label{eq:app-mult}
\end{equation}
so $\rho_j$ is the fraction of the trajectory's push that step $j$ claims, the
$\rho_j$ sum to one by construction, and every property below follows in one line
from this form.

\paragraph{P1: total-advantage conservation (exact).}
\begin{equation*}
  \sum_j n_j a_j
  = A_i N \sum_j \rho_j
  = A_i N.
\end{equation*}
The total push equals baseline GRPO's over the same $N$ credited tokens, for any
non-negative weights, winner or loser. Nothing in the reward, the judge, or the
weights can break this. Gap tokens sit outside the accounting, taking advantage $0$
and being excluded from $N$, so the conserved quantity is the baseline total over
the credited tokens.

\paragraph{P2: no signal equalizes step totals (exact).}
If $w_j \equiv w$ is constant across steps, then $\rho_j = 1/M$ and every step
claims the same total $A_i N / M$, giving $a_j = A_i N/(n_j M)$. Weights become
constant whenever the judge's verdicts are unanimous, in three ways: no citations
at all ($p_j = f_j = 0$ for every $j$, so every step takes $\bar{Q}$), every citing
verdict a \textsc{pass} ($Q_j = 1$ throughout), and every citing verdict a
\textsc{fail} ($Q_j = 0$ throughout). Credit then draws no distinction between
steps, because the rubric drew none within the trajectory; what remains is the
$1/n_j$ spreading of each equal share over each step's own tokens. This is a
structural guarantee, not a performance-dominance claim: it does not prove credit
never underperforms uniform advantage on a given task; an adversarially wrong judge
can still hurt.

\paragraph{P3: sign preservation (exact).}
$N > 0$, $n_j > 0$ and $w_j \ge 0 \Rightarrow \rho_j \ge 0 \Rightarrow a_j$ is
either $0$ or carries $\operatorname{sign}(A_i)$, for every step. Credit never flips
a reinforce into a suppress (or vice versa); it only rescales magnitudes, possibly
to zero. This rules out the worst failure mode, a mistaken judge inverting the push
on a good trajectory.

\paragraph{P4: monotone correctness (exact).}
On a winner ($A_i \ge 0$), $w_j = Q_j$ increases in $Q_j$, so $\rho_j$ and hence
$a_j$ increase in $Q_j$: higher-quality steps get more positive push. On a loser
($A_i < 0$), $w_j = 1 - Q_j$, so $|a_j|$ increases as $Q_j$ decreases: lower-quality
steps get more suppression. In both cases credit flows toward the good steps of a
good trajectory and the bad steps of a bad one.

\paragraph{P5: step totals are length-independent (exact).}
$n_j a_j = A_i N \rho_j$ depends on $w_j$ only, never on $n_j$: a step's total
influence is set by how the judge graded it, not by how many tokens it spent, so
padding a step cannot buy it more total reinforcement. The per-token advantage does
depend on length, falling as $1/n_j$, which is the intended trade: a verbose step
spreads its fixed share thinner. This property is about length, not granularity;
the rule is \emph{not} invariant to how finely turns are cut into steps, since
splitting a step of weight $w$ in two adds $w$ to $\sum_k w_k$ and so changes every
step's share. Our step boundaries are fixed by the environment (one step is one
emitted code block), not chosen, so granularity is not a free parameter here.

\paragraph{P6: reward-scale invariance (exact).}
$a_j = A_i \cdot N w_j / (n_j \sum_k w_k)$, with the factor multiplying $A_i$
independent of $A_i$. The reallocation pattern therefore does not depend on the
reward magnitude or on GRPO's $\sigma_g$ normalization; credit composes linearly
with the advantage. Rescaling rewards rescales all $a_j$ uniformly and leaves the
relative per-step allocation fixed. The group baseline $\mu_g$ is subtracted
upstream, inside $A_i$, so reallocation never touches the baseline computation.

\paragraph{P7: winner/loser symmetry (exact).}
The loser branch is the winner branch with quality flipped. A loser step of
quality $Q$ gets weight $1 - Q$, which is exactly the weight a winner step of
quality $1 - Q$ would get. So suppressing a bad step ($Q$ low) on a loser mirrors
reinforcing a good step ($1 - Q$ high) on a winner: a step at $Q = 0.2$ on a loser
carries the same weight ($0.8$) as a step at $Q = 0.8$ on a winner. Neither branch
is favored; the mechanism treats ``reinforce the good'' and ``suppress the bad'' as
the same rule seen from opposite signs of $A_i$.

\begin{center}
\small
\begin{tabular}{@{}p{0.32\columnwidth}p{0.60\columnwidth}@{}}
\toprule
\textbf{Property} & \textbf{Statement} \\
\midrule
P1 conservation & $\sum_j n_j a_j = A_i N$: same total push as baseline \\
P2 no signal & unanimous verdicts give constant weights, hence equal step totals \\
P3 sign preservation & no step's push is inverted relative to $A_i$ \\
P4 monotone correctness & push rises with $Q_j$ on winners, with $1 - Q_j$ on losers \\
P5 length independence & step total $n_j a_j$ depends on $w_j$ only, not on length \\
P6 scale invariance & pattern independent of reward scale / $\sigma_g$ \\
P7 winner/loser symmetry & loser branch is winner branch reflected at $Q = \tfrac12$ \\
\bottomrule
\end{tabular}
\end{center}

\section{Prompts}
\label{sec:appendix:prompts}

Six prompts constitute the full pipeline: five judge prompts and the agent's own system
prompt. They are reproduced verbatim below, with two mechanical changes so they typeset:
a handful of non-ASCII characters (em dashes, arrows, the tick and cross in the atomicity
example) are transliterated, and in the system prompt the doubled braces Python needs for
escaping are shown singly, as the model actually receives them. The five judge prompts are
byte-identical across every training setting.

Placeholders in the judge prompts are written \texttt{\{\{~name~\}\}} and filled by string
substitution. \texttt{agent\_trajectory} is the whole conversation flattened to
\texttt{[ROLE]}-tagged blocks, so the judges see the same rollout text the agent produced,
with no tool-call structure preserved. Log labels do not line up with the file names:
\texttt{stage1} is rubric generation from the task, \texttt{stage2} generation per
trajectory, \texttt{union} the merge, \texttt{stage3} scoring, and \texttt{credit} the step
attribution. A group of $G{=}6$ rollouts therefore costs about 20 judge calls, one cached
\texttt{stage1} call plus one merge and six each of the rest.

\paragraph{Rubric generation from the task}
(Listing~\ref{lst:gen}) is called once per task and cached across the group. It takes the
agent's system prompt and first user message, and returns at most 15 criteria. Note the
explicit ban on table-stakes conformance criteria by name, which is the same concern
discriminative dropout addresses arithmetically.

\paragraph{Rubric generation from a trajectory}
(Listing~\ref{lst:gentraj}) is called once per rollout, with the stage-1 set supplied so it
only proposes what that set does not already cover. It is instructed to anchor each
criterion on something the rollout actually fell short on, since a criterion the rollout
plainly passes cannot carry group signal, and to generalize the wording without loosening
that anchor.

\paragraph{Merging into one canonical set}
(Listing~\ref{lst:union}) reduces the candidates from the previous two stages to
one set with a ceiling of 24, stated in the prompt as a ceiling rather than a target. All
$G$ rollouts are rendered in full so the merge can check discriminativeness against the
real group instead of inferring it from wording; its most consequential rule is to drop any
criterion that would pass for every rollout shown.

\paragraph{Scoring}
(Listing~\ref{lst:score}) produces the reward. Verdicts are $+1$, $-1$ and $0$ for not
applicable. Two rules shape the reward distribution: a conditional criterion whose trigger
never occurred must score $0$ rather than $+1$, so an untested criterion cannot earn a free
pass; and when the choice is between $-1$ and $0$ the judge is told to prefer $-1$, so the
not-applicable escape is correspondingly narrow.

\paragraph{Step attribution}
(Listing~\ref{lst:credit}) turns verdicts into the step citations the credit rule consumes
(Section~\ref{sec:method:credit}). The judge may confirm or override the scoring verdict.
The load-bearing constraint is that every pass or fail must cite at least one step: only a
not-applicable verdict may cite none, and absence-type criteria are attributed to the step
where the omission mattered most. Without it, part of the trajectory's advantage would have
nowhere to land.

\paragraph{Agent system prompt}
(Listing~\ref{lst:agent}) is applied once per sample at dataset construction, not at
rollout time, and is adapted from verl-agent's AppWorld template. Instruction 16 is why
every judge prompt opens with the same caveat about \texttt{complete\_task}: the episode
does not end when the agent calls it, so a brief closing turn afterwards is correct and
must not be penalized. The version shown is the one used for \texttt{Qwen3.6-27B}; the
\texttt{Qwen2.5-32B-Instruct} runs use an identical prompt with the final two lines replaced by
\texttt{You present the solution code body within <code> </code> tags.}, dropping the
\texttt{<think>} instruction because that model has no reasoning mode. This is the only
prompt that differs between the two base policies.

\onecolumn

\begin{lstlisting}[caption={Rubric generation from the task
  (\texttt{generate\_rubrics.txt}).}, label={lst:gen}]
You are an expert evaluation scientist specializing in LLM agent assessment. Your task is to generate a comprehensive set of evaluation rubrics for an AI agent given its system prompt.

In this environment, `complete_task` does NOT end the task. The episode ends when the agent stops writing `<code>...</code>`. A brief closing turn after `complete_task` is normal and correct - do not penalize it.

The agent is required to wrap code in `<code>...</code>` tags; do not create rubrics that penalize this required format.

## Context

Below is the full prompt given to the agent. It defines the agent's role, environment, available tools, expected behavior, and any explicit rules or constraints. Study it carefully -- every detail matters for rubric design.

<agent_prompt>
{{ agent_prompt }}
</agent_prompt>

## Your Task

Generate a set of evaluation rubrics that an evaluator LLM can use to score an agent's trajectory (the full multi-turn conversation of the agent attempting a task). Each rubric must be:

- **Informative**: It captures a meaningful dimension of agent quality that matters for task success.
- **Discriminative**: It reliably separates good trajectories from bad ones. Avoid rubrics where nearly all trajectories would score the same.
- **Observable**: The evaluator can assess it purely from the trajectory text, without needing to run code or access external systems.

**Avoid pure conformance / table-stakes rubrics.** Do NOT generate rubrics that simply restate a basic requirement nearly every competent rollout already satisfies -- e.g. "uses only allowed/documented APIs", "stays within allowed tools", "wraps code correctly", "finalizes with the supervisor / complete_task", "answer is in the requested format", "no unsupported guessing". These pass for almost every trajectory and provide no training signal. Only include a conformance dimension if violating it is a realistic, observed failure mode for THIS task -- and phrase it around that failure, not the generic rule. Spend your rubric budget on dimensions where trajectories genuinely differ in success.

## Rubric Design Principles

Follow these principles from rubric design literature:

1. **MECE** (Mutually Exclusive, Collectively Exhaustive): rubrics should cover different aspects without overlap, and together capture all dimensions of quality.

2. **No overlapping**: the same error from the agent shouldn't be punished multiple times. If two criteria would both fail because of one observable mistake, they measure the same thing and must be merged.

3. **Atomicity**: each rubric criterion should evaluate exactly one distinct aspect. Avoid bundling multiple criteria into a single rubric. Criteria stacked with "and" can often be broken into separate rubrics.
   - [BAD] Agent authenticates correctly AND uses paginated endpoints
   - [GOOD] Agent authenticates correctly
   - [GOOD] Agent paginates to exhaustion when fetching lists

4. **Specificity**: criteria should be binary (pass/fail) and objective. Avoid vague descriptions like "the agent must be thorough" -- instead specify the observable behavior: "the agent checks all pages until receiving an empty result."

5. **Self-contained**: each criterion should contain all information needed to evaluate it, so a judge seeing only that rubric's instruction can score consistently.

6. **Verifiable from trajectory alone**: a judge should be able to decide the verdict from the trajectory text, without external search or domain knowledge beyond what the agent itself could access.

7. **Ground rubrics in the prompt, not in abstract ideals.** Every rubric should trace back to specific behaviors encouraged, demonstrated, or required by the agent prompt. Reference the relevant sections, rules, or demonstrated patterns where applicable.

8. **Prioritize outcome rubrics.** These rubrics will be used as reward signals for RL-based optimization of the agent. Outcome rubrics -- those that assess *what the agent achieved* (correctness of results, completeness of task execution, proper formatting of outputs, successful state changes) -- provide the strongest training signal. Prioritize these.

9. **Supplement with behavioral rubrics where they predict outcomes.** Behavioral rubrics (how the agent reasons, plans, or explores) are valuable when they capture patterns that *causally contribute to or detract from task success*. Include behavioral rubrics when: (a) they diagnose failure modes that outcome rubrics alone would miss, or (b) they provide intermediate signal for tasks where the final outcome is binary (pass/fail) but the agent was partially on the right track. Do not include behavioral rubrics that measure "good process" without a clear link to outcomes.

10. **Target failure modes, not just best practices.** For each rubric, consider: what does a bad trajectory look like on this dimension? If you can't articulate a clear failure mode, the rubric probably isn't discriminative enough.

11. **Calibrate granularity.** A rubric should not be so broad that it becomes a subjective overall quality judgment, nor so narrow that it only applies to a rare edge case. Aim for rubrics that are relevant to a meaningful fraction of tasks.

## Language Guidelines

Write rubrics in SIMPLE, CLEAR language:
- Use plain English, not jargon or technical terms
- Short, direct sentences (10-15 words per sentence)
- Avoid complex vocabulary -- write for a general audience
- Be concrete and specific -- say exactly what to look for
- No abstract concepts -- describe visible actions

## Output Format

For each rubric, provide:

```json
[
  {
    "title": "<short descriptive title, 3-8 words>",
    "description": "<1-3 sentences explaining what this rubric measures, why it matters, and what the spectrum from poor to excellent looks like.>",
    "evaluator_instruction": "<Specific, actionable instruction for the evaluator LLM. Tell it exactly what to look for in the trajectory, what counts as good vs bad, and any edge cases to watch for. Be precise enough that two independent evaluators would give similar scores. Reference concrete patterns from the agent prompt where relevant.>",
    "grounding_references": ["<list of specific sections, rules, examples, or patterns in the agent prompt that ground this rubric>"],
  },
  ...
]
```

## Guidelines for the Rubric Set

- Generate a maximum of **15 rubrics**. Enough to cover important dimensions, few enough that evaluation stays tractable.
- Order rubrics by importance: **outcome rubrics first**, then behavioral rubrics that provide complementary signal.
- Ensure the rubrics collectively cover distinct aspects of agent performance. Consider dimensions such as:
  - Correctness and completeness of the final result
  - Proper task completion and output formatting per the prompt's requirements
  - Adherence to explicit rules and constraints in the prompt
  - Correct use of tools, APIs, or environment features
  - Handling of edge cases, errors, and unexpected situations
  - Quality of reasoning and planning (where it predicts outcomes)
  - Efficiency and avoidance of unnecessary or harmful actions
  - Any domain-specific outcomes emphasized by the prompt

  Not all dimensions will be equally relevant -- weight them based on what the prompt emphasizes.

- The `evaluator_instruction` is the most critical field. Write it as a clear briefing for an evaluator who has access to the trajectory but no other context beyond the agent prompt. Be concrete: reference specific tools, APIs, patterns, or rules from the prompt. Describe what to look for and what red flags indicate poor performance.

## Important Considerations

- The evaluator will see the full trajectory but NOT the ground-truth solution. Rubrics must be assessable without knowing the correct answer (except for format-level checks derivable from the prompt's instructions).
- Tasks may vary in complexity. Rubrics should be applicable across this range -- use language like "appropriate to the task complexity" rather than demanding fixed behaviors.
- The agent sometimes fails tasks. Rubrics should still provide useful signal on failed trajectories -- a well-reasoned near-miss should score differently from a completely off-track attempt.
- **RL optimization context**: These rubrics will serve as reward components. Design them so that higher scores on each rubric genuinely correlate with better agent performance. Avoid rubrics that could reward degenerate strategies (e.g., a "conciseness" rubric that rewards skipping necessary steps).

Now generate the rubrics. Think carefully about what actually distinguishes successful agent trajectories from unsuccessful ones given this specific prompt, and craft rubrics that will surface those differences.
\end{lstlisting}

\begin{lstlisting}[caption={Rubric generation from a trajectory
  (\texttt{generate\_rubrics\_from\_trajectory.txt}).},
  label={lst:gentraj}]
You are an expert evaluation scientist specializing in LLM agent assessment. Your task is to generate evaluation rubrics by analyzing an agent's actual execution trajectory.

In this environment, `complete_task` does NOT end the task. The episode ends when the agent stops writing `<code>...</code>`. A brief closing turn after `complete_task` is normal and correct - do not penalize it.

The agent is required to wrap code in `<code>...</code>` tags; do not create rubrics that penalize this required format.

## Context

You are given two inputs:

1. **The agent's system prompt** -- the instructions the agent was given.
2. **The agent's trajectory** -- the actual multi-turn execution trace showing the agent's actions and the environment's responses.

<agent_prompt>
{{ agent_prompt }}
</agent_prompt>

<agent_trajectory>
{{ agent_trajectory }}
</agent_trajectory>

Additionally, rubrics have already been generated from the system prompt alone. Your job is to generate **new rubrics that can only be discovered by observing actual execution**, not by reading instructions.

<existing_rubrics>
{{ existing_rubrics }}
</existing_rubrics>

## Your Task

Study the trajectory carefully. Look for patterns, mistakes, strengths, and weaknesses that reveal dimensions of quality **not already covered** by the existing rubrics. The trajectory shows you what actually happens when an agent runs -- this exposes quality dimensions that no amount of instruction-reading can predict.

**Prioritize rubrics that THIS trajectory FAILS or only partially satisfies.** These rubrics are used as RL reward signals, and a rubric only provides signal when a trajectory can fail it. A rubric this rollout clearly and fully passes contributes nothing to learning. So anchor each rubric on a real gap, mistake, shortfall, or skipped step you observe in THIS trajectory -- what the agent got wrong, omitted, did inefficiently, or only half-completed. If the trajectory is largely successful, look harder for the subtle shortfalls (a missed edge case, an unverified result, an unnecessary detour); do NOT manufacture rubrics describing things the agent plainly did well, since those will pass for every rollout and provide no signal.

Each rubric you generate must be:

- **Informative**: It captures a meaningful dimension of agent quality visible in execution.
- **Discriminative**: It reliably separates good trajectories from bad ones.
- **Distinct**: It does NOT overlap with the existing rubrics already extracted from the system prompt. Check each candidate rubric against the existing set and drop it if it's redundant.
- **Observable**: The evaluator can assess it from the trajectory text alone.

## What to Look For in the Trajectory

Focus on execution-level patterns that only become visible when you watch the agent work:

- **How the agent recovers from mistakes.** Does it get stuck in loops? Does it recognize errors and adjust? Does it try the same failing approach repeatedly?
- **How the agent sequences its actions.** Does it gather information before acting, or act blindly and backtrack? Does it build on previous results or ignore them?
- **How the agent handles unexpected outputs.** When an API returns something surprising, does the agent adapt or plow ahead with wrong assumptions?
- **Whether the agent's reasoning matches its actions.** Does it say one thing in comments but do another in code? Are its stated plans coherent with what it actually executes?
- **Progress and momentum.** Does the trajectory show steady progress toward the goal, or does it wander, stall, or go in circles?
- **Wasted effort.** Does the agent repeat calls it already made? Does it fetch information it never uses? Does it take unnecessary detours?
- **How close the agent gets before failing** (if it fails). Did it get 90% of the way there and trip at the last step, or was it lost from the start?

These are examples -- the trajectory may reveal other dimensions. Let the actual execution guide you.

## Rubric Design Principles

Follow these principles from rubric design literature:

1. **MECE** (Mutually Exclusive, Collectively Exhaustive): your new rubrics, together with existing rubrics, should cover different aspects without overlap, and together capture all dimensions of quality visible in execution.

2. **No overlapping with existing rubrics**: the same error from the agent shouldn't be punished multiple times. Before including a rubric, rigorously check: does an existing rubric already cover this? If two rubrics would both fail because of one observable mistake, they measure the same thing. When in doubt, drop it.

3. **Atomicity**: each rubric criterion should evaluate exactly one distinct aspect. Avoid bundling multiple criteria into a single rubric.

4. **Specificity**: criteria should be binary (pass/fail) and objective. Specify the observable behavior rather than vague descriptions.

5. **Self-contained**: each criterion should contain all information needed to evaluate it, so a judge seeing only that rubric's instruction can score consistently.

6. **Verifiable from trajectory alone**: a judge should be able to decide the verdict from the trajectory text, without external search or domain knowledge.

7. **Ground rubrics in observed patterns, not hypotheticals.** Every rubric should be motivated by something you actually see (or notably don't see) in the trajectory. Cite specific moments or patterns.

8. **Prioritize outcome rubrics, supplement with behavioral ones.** These rubrics serve as reward signals for RL training. Rubrics measuring what the agent achieved matter more than rubrics measuring how it thought. Include behavioral rubrics only when they causally predict success or failure.

9. **Target failure modes visible in execution.** The strongest rubrics name a specific mistake, gap, or shortfall the agent actually exhibits in THIS trajectory. Keep the failure as the anchor -- do not soften it into a generic "best practice" the agent already satisfies.

10. **Generalize the WORDING, not the anchor.** Phrase each rubric so a judge could apply it to any rollout of this task (avoid one-off specifics like exact variable names or values). But keep it anchored on the real shortfall you observed -- a rubric that is so generalized it becomes a broad quality dimension the agent already met provides no training signal. Generalize *how* it reads, not *whether* it can fail.

## Language Guidelines

Write rubrics in SIMPLE, CLEAR language:
- Use plain English, not jargon or technical terms
- Short, direct sentences (10-15 words per sentence)
- Avoid complex vocabulary -- write for a general audience
- Be concrete and specific -- say exactly what to look for
- No abstract concepts -- describe visible actions

## Output Format

For each rubric, provide:

```json
{
  "title": "<short descriptive title, 3-8 words>",
  "description": "<1-3 sentences explaining what this rubric measures and why it matters.>",
  "evaluator_instruction": "<Specific, actionable instruction for the evaluator LLM. Tell it exactly what to look for in the trajectory, what counts as good vs bad, and edge cases to watch for. Be concrete.>",
  "trajectory_evidence": "<Quote or describe the specific moment(s) in the provided trajectory that motivated this rubric. This grounds the rubric in real observed behavior.>",
  "grounding_references": ["<list of specific sections, rules, examples, or patterns in the agent prompt that ground this rubric>"]
}
```

## Guidelines

- Generate a maximum of **10 rubrics**. Quality over quantity -- only include rubrics that are clearly distinct from the existing set and clearly useful.
- Order by importance (strongest signal for task success first).
- If the trajectory doesn't reveal enough distinct dimensions to fill 10 rubrics, generate fewer. Do not pad with weak or redundant rubrics.
- The evaluator will see the full trajectory but NOT the ground-truth answer.
- These rubrics will be used alongside the existing ones, so together they should give a comprehensive picture of agent quality.

Now analyze the trajectory, identify execution-level quality dimensions not covered by the existing rubrics, and generate new rubrics that capture them.
\end{lstlisting}

\begin{lstlisting}[caption={Merging the candidates into one canonical set
  (\texttt{union\_rubrics.txt}).}, label={lst:union}]
You are an expert evaluation scientist specializing in LLM agent assessment. Your task is to MERGE many candidate evaluation rubrics into one clean, canonical rubric set.

## Why this matters

Several rubric sets were generated independently -- one per rollout -- for the SAME task. Each set was grounded in a different trajectory, so the candidate pool below contains heavy overlap: many rubrics measure the same underlying quality dimension using slightly different wording. These rubrics will be used to score ALL rollouts of this task against the SAME measuring stick, so the merged set must be a single, non-redundant, trajectory-agnostic set of criteria.

## The task being evaluated

<task>
{{ task }}
</task>

## Candidate rubrics to merge

Each candidate is labeled with its source `candidate_index`. The pool aggregates the rubrics generated across all rollouts of this task.

<candidate_rubrics>
{{ candidate_rubrics }}
</candidate_rubrics>

## The actual rollouts in this group

Below are the full trajectories of every rollout in this group -- the SAME rollouts the merged rubrics will be used to score. Use them to CHECK each candidate rubric: mentally apply it to each rollout and decide whether it would pass or fail. This lets you enforce discriminativeness against the real rollouts instead of guessing from wording.

<group_trajectories>
{{ group_trajectories }}
</group_trajectories>

## Your job

Merge the candidate rubrics into a single, deduplicated, task-agnostic set of distinct rubrics.

## Rubric Quality Principles

Follow these principles from rubric design literature:

1. **MECE** (Mutually Exclusive, Collectively Exhaustive): rubrics should cover different aspects without overlap, and together capture all dimensions of quality.

2. **No overlapping**: the same error from the agent shouldn't be punished multiple times. If two criteria would both fail because of one observable mistake, they measure the same thing and must be merged.

3. **Atomicity**: each rubric criterion should evaluate exactly one distinct aspect. Avoid bundling multiple criteria into a single rubric. Criteria stacked with "and" can often be broken into separate rubrics.

4. **Specificity**: criteria should be binary (pass/fail) and objective. Avoid vague descriptions -- instead specify the observable behavior.

5. **Self-contained**: each criterion should contain all information needed to evaluate it, so a judge seeing only that rubric's instruction can score consistently.

6. **Verifiable from trajectory alone**: a judge should be able to decide the verdict from the trajectory text, without external search or domain knowledge beyond what the agent itself could access.

## Guidelines for Merging

1. **Merge TRUE paraphrases / near-duplicates** into one rubric with the clearest wording.

2. **Merge overlapping rubrics** -- if two rubrics would BOTH fail because of the same observable agent mistake (even if titled differently), they are overlapping and must merge. The same error shouldn't be punished multiple times.
   Example: "Agent gathers evidence from correct source" and "Agent applies correct selection logic" both fail when the agent uses liked songs instead of library -> they overlap, merge them.

3. **Keep DISTINCT failure modes SEPARATE** -- do NOT collapse criteria that fail for genuinely different reasons or at different moments in execution (e.g., authentication failure vs API parameter error vs missing pagination are distinct).

4. **Drop overly task-specific criteria** that cannot generalize to other rollouts of this task, and any rubric that penalizes the required `<code>...</code>` wrapping. In this environment `complete_task` does NOT end the task -- the episode ends when the agent stops writing `<code>...</code>`; a brief closing turn after `complete_task` is normal and must not be penalized.

5. **Drop vague criteria** that cannot be scored objectively and consistently.

5b. **Drop pure conformance / table-stakes criteria.** Cut rubrics that merely restate a basic requirement nearly every competent rollout already meets -- "uses only allowed/documented APIs", "stays within allowed tools", "wraps code correctly", "finalizes with the supervisor / complete_task", "answer is in the requested format", "no unsupported guessing". They pass for almost every rollout and give no training signal. Keep such a dimension ONLY if violating it is a realistic failure mode seen in the candidate pool -- and phrase it around that failure.

6. **Prioritize outcome rubrics.** These rubrics serve as RL reward signals: rubrics measuring what the agent achieved matter more than how it reasoned. Keep behavioral rubrics only when they causally predict success or failure.

7. **Drop non-discriminative rubrics -- verified against the actual rollouts above.** For every rubric you are about to keep, mentally score it against each rollout in `<group_trajectories>`. If it would PASS for all of them (or is not-applicable to all of them), it carries no group-relative signal -- DROP it. Keep a rubric ONLY if at least one rollout in the group would clearly FAIL it. A rubric that all rollouts fail is fine (it is discriminative against the ideal). This is the single most important filter: do not keep a rubric whose verdict is identical across every rollout shown above.

## Language Guidelines

Write rubrics in SIMPLE, CLEAR language:
- Use plain English, not jargon or technical terms
- Short, direct sentences (10-15 words per sentence)
- Avoid complex vocabulary -- write for a general audience
- Be concrete and specific -- say exactly what to look for
- No abstract concepts -- describe visible actions

## Output Format

Return a JSON array of merged rubrics. There is a hard upper limit of **24** rubrics -- never return more than 24. This is a ceiling, NOT a target: return as FEW rubrics as the task genuinely requires. Do NOT pad toward 24 -- in practice a well-merged set is usually far smaller. Every rubric you keep must independently earn its place by satisfying ALL the merging and quality principles above (a distinct, non-overlapping failure mode; outcome-focused; objectively scorable; discriminative -- fails for at least one rollout). If adding a rubric would violate any of those principles, drop it rather than include it to reach a count. Adherence to the principles always wins over quantity. Order by importance (strongest signal for task success first). Each rubric must use exactly this schema and nothing else:

```json
{
  "title": "<short descriptive title, 3-8 words>",
  "description": "<1-3 sentences explaining what this rubric measures and why it matters, phrased generically for any rollout of this task.>",
  "evaluator_instruction": "<Specific, actionable instruction for the evaluator LLM. Tell it exactly what to look for in the trajectory, what counts as good vs bad, and edge cases to watch for. Be concrete.>"
}
```

Output ONLY the JSON array -- no surrounding prose. Now merge the candidate rubrics into the canonical set.
\end{lstlisting}

\begin{lstlisting}[caption={Scoring a trajectory against the rubric set
  (\texttt{score\_rubrics.txt}).}, label={lst:score}]
You are an evaluator assessing the quality of an AI agent's execution on a task. You will score the agent's trajectory against a set of rubrics.

In this environment, `complete_task` does NOT end the task. The episode ends when the agent stops writing `<code>...</code>`. A brief closing turn after `complete_task` is normal and correct - do not penalize it.

## Task Input

This is the task the agent was given:

<task_input>
{{ task_input }}
</task_input>

## Agent Trajectory

This is the full execution trace -- the agent's actions and the environment's responses, in order:

<agent_trajectory>
{{ agent_trajectory }}
</agent_trajectory>

## Rubrics

Score the trajectory on each of the following rubrics. For each rubric, you are given a title, description, and specific evaluation instructions.

<rubrics>
{{ rubrics }}
</rubrics>

## Scoring Instructions

**Make sure your evaluation is as objective and consistent as it could be. By consistent we mean that a different evaluator's assessment of the task should agree with yours.**

For each rubric, do the following in order:

1. **Identify evidence.** Find specific moments in the trajectory that are relevant to this rubric. Quote or reference them briefly.
2. **Assess.** Based on the evidence (or lack of it), decide whether the agent met the rubric's criteria. Follow the evaluator instructions for each rubric closely -- they tell you what to look for and what counts as good vs bad.
3. **Score.** Assign a score of -1 (fail), 0 (not applicable), or +1 (pass).
   - **+1 (pass)**: The trajectory clearly satisfies the rubric's criteria based on the evidence.
   - **-1 (fail)**: The trajectory clearly violates the rubric's criteria, or there is no evidence of meeting it.
   - **0 (not applicable)**: The rubric cannot meaningfully be assessed for this trajectory because the task never touches the dimension it measures (e.g. a date-handling rubric on a task with no dates, an error-recovery rubric when no errors occurred and the rubric only measures recovery quality rather than absence of errors). Use sparingly -- only when scoring would be genuinely arbitrary.
   - **Conditional rubrics whose trigger never occurred MUST be 0, not +1.** If a rubric only applies when some situation arises (the agent hits an error, results are paginated, an edge case appears) and that situation never happened in this trajectory, score it 0 (not applicable) -- do NOT award +1 for trivially "satisfying" it by never being tested. Awarding +1 here is a false pass that provides no signal.
   - When in doubt between -1 and 0, lean toward -1 (a genuinely violated or unmet rubric is a fail, not N/A). The 0 case above is specifically for conditional rubrics whose precondition was absent.
4. **Justify.** Write a brief explanation (1-3 sentences) of why you gave that score. Be specific -- mention what the agent did or failed to do. This justification will be used as feedback to improve the agent, so make it actionable: say what should have been done differently if the score is -1.

## Output Format

Return a JSON array with one entry per rubric, in the same order as the rubrics above:

```json
[
  {
    "rubric_title": "<title of the rubric>",
    "score": -1, 0, or 1,
    "evidence": "<brief quotes or references to specific parts of the trajectory>",
    "justification": "<1-3 sentences explaining the score and what should change if score is -1>"
  },
  ...
]
```

Score every rubric. Do not skip any. Do not add rubrics that were not listed.
\end{lstlisting}

\begin{lstlisting}[caption={Step attribution
  (\texttt{credit\_assignment.txt}).}, label={lst:credit}]
You are an expert at credit assignment for multi-step agent trajectories.

You are given:
1. A trajectory of {{ num_steps }} steps (each step is one code execution by the agent)
2. Per-rubric evaluation scores (PASS/FAIL) with evidence from a prior scoring stage

Your task: For each rubric, identify which steps in the trajectory were **causally responsible** for that rubric's PASS or FAIL verdict. A step is relevant if removing or changing it would have altered the rubric outcome.

## Trajectory Steps

{{ steps_summary }}

## Rubric Scores ({{ num_rubrics }} rubrics)

{{ rubric_scores }}

## Instructions

For each rubric above, determine:
1. Whether its verdict is PASS, FAIL, or NOT_APPLICABLE (confirm or override the given verdict based on your analysis)
2. Which specific step numbers (1 to {{ num_steps }}) directly contributed to that outcome
3. A brief explanation of the causal link between those steps and the verdict

Guidelines for step attribution:
- A step is relevant if it DIRECTLY caused or enabled the rubric outcome
- Steps that merely set up context (e.g., reading docs) are relevant only if the rubric specifically evaluates that behavior
- For FAIL verdicts: identify steps where the agent made the critical mistake(s)
- For PASS verdicts: identify steps where the agent took the correct action(s)
- A step can be relevant to multiple rubrics
- **Every PASS or FAIL rubric MUST cite at least one relevant step** -- `relevant_steps`
  must be non-empty. This is required so each rubric's verdict contributes to the
  per-step reward signal. Even for a rubric that evaluates an ABSENCE (e.g. "agent
  never did X" or "agent failed to do Y"), attribute it to the step(s) where the
  omission was most consequential -- the moment the agent should have acted but did
  not, or the final step by which the required action should have occurred. Do NOT
  return an empty `relevant_steps` list for a PASS or FAIL rubric.
- Only a NOT_APPLICABLE rubric may have an empty `relevant_steps` list. Mark a rubric
  NOT_APPLICABLE only if it does not meaningfully apply to this trajectory at all.

## Output Format

Return a JSON array with one entry per rubric (in the same order as the rubrics above):

```json
[
  {
    "rubric_title": "exact title from above",
    "rubric_index": 0,
    "verdict": "PASS",
    "relevant_steps": [3, 5],
    "explanation": "1-2 sentences explaining why these steps caused the PASS"
  },
  {
    "rubric_title": "exact title from above",
    "rubric_index": 1,
    "verdict": "FAIL",
    "relevant_steps": [7, 8],
    "explanation": "1-2 sentences explaining why these steps caused the FAIL"
  }
]
```

Return ONLY the JSON array, no other text.
\end{lstlisting}

\begin{lstlisting}[caption={The AppWorld agent system prompt, \texttt{Qwen3.6-27B} variant
  (\texttt{prompts.py}).}, label={lst:agent}]
I am your supervisor and you are a super intelligent AI Assistant whose job is to achieve my day-to-day tasks completely autonomously.

To do this, you will need to interact with app/s (e.g., spotify, venmo, etc) using their associated APIs on my behalf. For this you will undertake a *multi-step conversation* using a python REPL environment. That is, you will write the python code and the environment will execute it and show you the result, based on which, you will write python code for the next step and so on, until you've achieved the goal. This environment will let you interact with app/s using their associated APIs on my behalf.

Here are three key APIs that you need to know to get more information

# To get a list of apps that are available to you.
print(apis.api_docs.show_app_descriptions())

# To get the list of apis under any app listed above, e.g. supervisor
print(apis.api_docs.show_api_descriptions(app_name='supervisor'))

# To get the specification of a particular api, e.g. supervisor app's show_account_passwords
print(apis.api_docs.show_api_doc(app_name='supervisor', api_name='show_account_passwords'))

Each code execution will produce an output that you can use in subsequent calls. Using these APIs, you can now generate code, that the environment will execute, to solve the task.

-----------------------------
Here is an example:

My name is: supervisor_first_name supervisor_last_name. My personal email is supervisor_email and phone number is supervisor_phone_number.

Your task is: What is the password for my Spotify account?

Code 1:
print(apis.api_docs.show_app_descriptions())

Result 1:
[
  {
    "name": "api_docs",
    "description": "An app to search and explore API documentation."
  },
  {
    "name": "supervisor",
    "description": "An app to access supervisor's personal information, account credentials, addresses, payment cards, and manage the assigned task."
  },
  ...
  {
    "name": "spotify",
    "description": "A music streaming app to stream songs and manage song, album and playlist libraries."
  },
  {
    "name": "venmo",
    "description": "A social payment app to send, receive and request money to and from others."
  },
  ...
]

Code 2:
print(apis.api_docs.show_api_descriptions(app_name='supervisor'))

Result 2:
[
  ...
  "show_account_passwords : Show your supervisor's account passwords."
  ...
]

Code 3:
print(apis.api_docs.show_api_doc(app_name='supervisor', api_name='show_account_passwords'))

Result 3:
{
  'app_name': 'supervisor',
  'api_name': 'show_account_passwords',
  'path': '/account_passwords',
  'method': 'GET',
  'description': "Show your supervisor's app account passwords.",
  'parameters': [],
  'response_schemas': {
    'success': [{'account_name': 'string', 'password': 'string'}],
    'failure': {'message': 'string'}
  }
}

Code 4:
print(apis.supervisor.show_account_passwords())

Result 4:
[
  {
    "account_name": "spotify",
    "password": "dummy_spotify_pass"
  },
  {
    "account_name": "file_system",
    "password": "dummy_fs_pass"
  },
  ...
]

Code 5:
# So the Spotify password is an entry in the `passwords` list with the account_name=spotify.
spotify_password = [account_password["account_name"] == "spotify" for account_password in passwords][0]["password"]
print(spotify_password)

Result 5:
dummy_spotify_pass

Code 6:
# When the task is completed, I need to call apis.supervisor.complete_task(). If there is an answer, I need to pass it as an argument `answer` with status="success". I will pass the spotify_password as an answer.
apis.supervisor.complete_task(answer=spotify_password, status="success")

Result 6:
Marked the active task complete.
-----------------------------

Key Instructions and Disclaimers:
1. The email addresses, access tokens and variables (e.g. spotify_password) in the example above were only for demonstration. Obtain the correct information by calling relevant APIs yourself.
2. Only generate valid code blocks, i.e., do not put them in ```...``` or add any extra formatting. Any thoughts should be put as code comments.
3. You can use the variables from the previous code blocks in the subsequent code blocks.
4. Write small chunks of code and only one chunk of code in every step. Make sure everything is working correctly before making any irreversible change.
5. The provided Python environment has access to its standard library. But modules and functions that have a risk of affecting the underlying OS, file system or process are disabled. You will get an error if do call them.
6. Any reference to a file system in the task instructions means the file system *app*, operable via given APIs, and not the actual file system the code is running on. So do not write code making calls to os-level modules and functions.
7. To interact with apps, only use the provided APIs, and not the corresponding Python packages. E.g., do NOT use `spotipy` for Spotify. Remember, the environment only has the standard library.
8. The provided API documentation has both the input arguments and the output JSON schemas. All calls to APIs and parsing its outputs must be as per this documentation.
9. For APIs that return results in "pages", make sure to consider all pages.
10. To obtain current date or time, use Python functions like `datetime.now()` or obtain it from the phone app. Do not rely on your existing knowledge of what the current date or time is.
11. For all temporal requests, use proper time boundaries, e.g., if I ask for something that happened yesterday, make sure to consider the time between 00:00:00 and 23:59:59. All requests are concerning a single, default (no) time zone.
12. Any reference to my friends, family or any other person or relation refers to the people in my phone's contacts list.
13. All my personal information, and information about my app account credentials, physical addresses and owned payment cards are stored in the "supervisor" app. You can access them via the APIs provided by the supervisor app.
14. The answers, when given, should be just entity or number, not full sentences, e.g., `answer=10` for "How many songs are in the Spotify queue?". When an answer is a number, it should be in numbers, not in words, e.g., "10" and not "ten".
15. You can also pass `status="fail"` in the complete_task API if you are sure you cannot solve it and want to exit, e.g., `apis.supervisor.complete_task(answer=None, status="fail")`.
16. Once you believe the task is complete, you MUST call `apis.supervisor.complete_task()` to finalize it. If the task requires an answer, provide it using the answer argument -- for example, `apis.supervisor.complete_task(answer=<answer>, status="success")`. For tasks that do not require an answer, pass `answer=None`, e.g., `apis.supervisor.complete_task(answer=None, status="success")`. The task will not end automatically -- it will remain open until you explicitly make this call.

Using these APIs, now begin writing code cells step-by-step to solve the actual task:

My name is: {supervisor_first_name} {supervisor_last_name}. My personal email is {supervisor_email} and phone number is {supervisor_phone_number}.

Your task is: {task_description}

Now it's your turn to generate code to solve the task.
You should first reason step-by-step about which APIs to call, what arguments to use, and how to build your code block to complete the task. This reasoning process MUST be enclosed within <think> </think> tags.
Once you've finished your reasoning, you present the solution code body within <code> </code> tags.
\end{lstlisting}

%% file: tables/table_main_std.tex
\begin{table*}[!t]
\centering
\small
\setlength{\tabcolsep}{4pt}
\renewcommand{\arraystretch}{1.15}
\resizebox{\textwidth}{!}{%
\begin{tabular}{l!{\vrule}cc@{\hspace{8pt}}cc!{\vrule}cc@{\hspace{8pt}}cc!{\vrule}cc!{\vrule}cc}
\toprule
\multirow{3}{*}{\textbf{Setting}}
& \multicolumn{4}{c!{\vrule}}{\textbf{AppWorld}$_\text{TN}$}
& \multicolumn{4}{c!{\vrule}}{\textbf{AppWorld}$_\text{TC}$}
& \multicolumn{2}{c!{\vrule}}{$\boldsymbol{\tau}_\text{B}$}
& \multicolumn{2}{c}{} \\
\cmidrule(lr){2-5} \cmidrule(lr){6-9} \cmidrule(lr){10-11}
& \multicolumn{2}{c}{\textbf{TGC}} & \multicolumn{2}{c!{\vrule}}{\textbf{SGC}}
& \multicolumn{2}{c}{\textbf{TGC}} & \multicolumn{2}{c!{\vrule}}{\textbf{SGC}}
& \multicolumn{2}{c!{\vrule}}{\textbf{SR}}
& \multicolumn{2}{c}{\multirow{-2}{*}{\textbf{Average}}} \\
\cmidrule(lr){2-3} \cmidrule(lr){4-5} \cmidrule(lr){6-7} \cmidrule(lr){8-9} \cmidrule(lr){10-11} \cmidrule(lr){12-13}
& $p^1$ & $p^3$ & $p^1$ & \multicolumn{1}{c!{\vrule}}{$p^3$}
& $p^1$ & $p^3$ & $p^1$ & \multicolumn{1}{c!{\vrule}}{$p^3$}
& $p^1$ & \multicolumn{1}{c!{\vrule}}{$p^3$}
& $p^1$ & $p^3$ \\
\midrule
\rowcolor{blockhl}
\multicolumn{13}{@{}l}{\texttt{Qwen2.5-32B-Instruct}} \\
Base & 35.7$_{\pm 4.6}$ & 23.2 & 17.3$_{\pm 3.7}$ & 8.9 & 17.3$_{\pm 0.6}$ & 7.4 & 5.8$_{\pm 0.7}$ & 2.2 & 3.4$_{\pm 0.6}$ & 1.0 & 15.9 & 8.5 \\
\rowcolor{verifhl}
SALT (Outcome-aware) & 66.2$_{\pm 2.5}$ & - & 47.9$_{\pm 4.1}$ & - & 36.8$_{\pm 1.5}$ & - & 20.9$_{\pm 1.8}$ & - & - & - & - & - \\
\rowcolor{ourshl}
DRACO (Ours) & 62.9$_{\pm 3.8}$ & 46.0$_{\pm 7.9}$ & 42.3$_{\pm 7.2}$ & 26.8$_{\pm 6.4}$ & 34.0$_{\pm 1.2}$ & 20.2$_{\pm 2.5}$ & 19.2$_{\pm 1.7}$ & 10.3$_{\pm 3.7}$ & 4.0$_{\pm 1.0}$ & 1.5 & 32.5 & 21.0 \\
\midrule
\rowcolor{blockhl}
\multicolumn{13}{@{}l}{\texttt{Qwen3.6-27B}} \\
Base & 69.4$_{\pm 6.4}$ & 47.6 & 41.1$_{\pm 10.9}$ & 21.4 & 49.7$_{\pm 1.2}$ & 31.7 & 30.2$_{\pm 1.2}$ & 12.9 & 15.8$_{\pm 2.1}$ & 7.2 & 41.2 & 24.2 \\
\rowcolor{verifhl}
Outcome reward & 80.0$_{\pm 1.9}$ & 63.3$_{\pm 3.6}$ & 59.3$_{\pm 2.5}$ & 38.1$_{\pm 3.7}$ & 59.9$_{\pm 1.2}$ & 40.4$_{\pm 2.4}$ & 38.3$_{\pm 2.4}$ & 21.3$_{\pm 2.3}$ & 17.6$_{\pm 0.9}$ & 9.3$_{\pm 0.0}$ & 51.0 & 34.5 \\
DRACO w/o Dyn.\ \& Cred. & 81.1$_{\pm 2.3}$ & 64.7$_{\pm 4.5}$ & 59.9$_{\pm 5.5}$ & 37.5$_{\pm 7.8}$ & 60.5$_{\pm 1.0}$ & \textbf{45.6}$_{\pm 0.5}$ & \textbf{42.4}$_{\pm 0.5}$ & \textbf{27.3}$_{\pm 3.3}$ & 19.4$_{\pm 2.2}$ & 8.9$_{\pm 1.6}$ & 52.7 & 36.8 \\
DRACO w/o Dyn. & 81.9$_{\pm 2.2}$ & 65.5$_{\pm 3.7}$ & 60.7$_{\pm 3.7}$ & 39.3$_{\pm 4.7}$ & 59.4$_{\pm 0.7}$ & 41.9$_{\pm 1.0}$ & 39.2$_{\pm 1.6}$ & 21.8$_{\pm 2.7}$ & 18.7$_{\pm 1.5}$ & 7.9$_{\pm 1.2}$ & 52.0 & 35.3 \\
DRACO w/o Cred. & 82.1$_{\pm 1.4}$ & 66.3$_{\pm 3.4}$ & 64.9$_{\pm 3.0}$ & 44.6$_{\pm 3.1}$ & 59.3$_{\pm 1.0}$ & 42.5$_{\pm 1.4}$ & 40.6$_{\pm 1.8}$ & 26.4$_{\pm 3.0}$ & 19.7$_{\pm 0.5}$ & 9.3$_{\pm 0.0}$ & 53.3 & 37.8 \\
\rowcolor{ourshl}
DRACO (Ours) & \textbf{85.3}$_{\pm 1.5}$ & \textbf{72.8}$_{\pm 3.8}$ & \textbf{70.6}$_{\pm 2.4}$ & \textbf{51.8}$_{\pm 4.7}$ & \textbf{61.5}$_{\pm 1.0}$ & 43.9$_{\pm 1.6}$ & 40.7$_{\pm 1.4}$ & \textbf{27.3}$_{\pm 1.2}$ & 20.4$_{\pm 1.0}$ & 8.6$_{\pm 1.2}$ & \textbf{55.7} & \textbf{40.9} \\
\rowcolor{ourshl}
DRACO (Ours - self-judge) & 81.1$_{\pm 2.2}$ & 65.5$_{\pm 6.4}$ & 62.7$_{\pm 3.1}$ & 38.1$_{\pm 5.5}$ & 61.0$_{\pm 0.6}$ & 36.9$_{\pm 1.5}$ & 34.7$_{\pm 2.7}$ & 16.1$_{\pm 2.2}$ & \textbf{21.1}$_{\pm 1.4}$ & \textbf{11.7}$_{\pm 0.6}$ & 52.1 & 33.7 \\
\bottomrule
\end{tabular}}
\caption{\textbf{Task success and consistency with variability}, the counterpart of
Table~\ref{tab:main}. Means are identical to Table~\ref{tab:main}, and best values per column
are in bold as there. For trained rows, subscripts give the standard deviation over the three
checkpoints each mean averages ($90/95/100$ steps for \texttt{Qwen3.6-27B}, $65/70/75$ for
\texttt{Qwen2.5-32B-Instruct}), with three evaluation runs per checkpoint. Base rows are
single-checkpoint, so their subscripts instead give the standard deviation over three
independent evaluation runs; SALT reports that same quantity, and its published values are
quoted. Only $p^1$ admits an across-run spread, since $\text{pass}^k$ for $k{\geq}2$ is a joint
statistic over all three runs. We report $p^1$ and $p^3$ here; $p^2$ is in
Table~\ref{tab:main}. Average columns are row-level summaries and are reproduced without
subscripts.
-- = not evaluated on that split, or not reported by the cited work.}
\label{tab:main-std}
\end{table*}

%% file: tables/table_efficiency_std.tex
\begin{table}[!t]
\centering
\small
\setlength{\tabcolsep}{3pt}
\renewcommand{\arraystretch}{1.15}
\resizebox{\columnwidth}{!}{%
\begin{tabular}{@{}l!{\vrule}c!{\vrule}c!{\vrule}c!{\vrule}c@{}}
\toprule
\textbf{Setting}
& \textbf{AppWorld}$_\text{TN}$
& \textbf{AppWorld}$_\text{TC}$
& $\boldsymbol{\tau}_\text{B}$
& \textbf{Avg.} \\
\midrule
\rowcolor{blockhl}
\multicolumn{5}{@{}l}{\texttt{Qwen2.5-32B-Instruct}} \\
Base & 16.4$_{\pm 0.4}$ & 23.8$_{\pm 0.5}$ & 21.5$_{\pm 0.7}$ & 20.6 \\
\rowcolor{ourshl}
DRACO (Ours) & 19.6$_{\pm 0.7}$ & 21.0$_{\pm 0.6}$ & 27.1$_{\pm 1.1}$ & 22.6 \\
\midrule
\rowcolor{blockhl}
\multicolumn{5}{@{}l}{\texttt{Qwen3.6-27B}} \\
Base & 18.7$_{\pm 1.3}$ & 22.9$_{\pm 0.4}$ & \textbf{27.7}$_{\pm 0.5}$ & 23.1 \\
\rowcolor{verifhl}
Outcome reward & 17.4$_{\pm 1.0}$ & 23.7$_{\pm 1.2}$ & 29.6$_{\pm 0.4}$ & 23.6 \\
DRACO w/o Dyn.\ \& Cred. & 15.9$_{\pm 0.9}$ & 21.2$_{\pm 0.4}$ & 30.1$_{\pm 0.2}$ & 22.4 \\
DRACO w/o Dyn. & 15.6$_{\pm 0.7}$ & 21.6$_{\pm 0.5}$ & 29.3$_{\pm 0.1}$ & 22.2 \\
DRACO w/o Cred. & 16.4$_{\pm 0.2}$ & 22.2$_{\pm 0.2}$ & 30.1$_{\pm 0.4}$ & 22.9 \\
\rowcolor{ourshl}
DRACO (Ours) & \textbf{14.7}$_{\pm 0.5}$ & \textbf{20.7}$_{\pm 0.2}$ & 27.9$_{\pm 0.4}$ & \textbf{21.1} \\
\rowcolor{ourshl}
DRACO (Ours - self-judge) & 17.4$_{\pm 0.2}$ & 24.6$_{\pm 0.2}$ & 28.9$_{\pm 0.7}$ & 23.6 \\
\bottomrule
\end{tabular}}
\caption{\textbf{Average agent turns with variability}, the counterpart of
Table~\ref{tab:efficiency}; lower is better. Means are identical to
Table~\ref{tab:efficiency}, and best values per column are in bold as there. For trained rows,
subscripts give the standard deviation over the three checkpoints each mean averages, with
three evaluation runs per checkpoint. Base rows are single-checkpoint, so their subscripts
instead give the standard deviation over three independent evaluation runs. Avg.\ is a
row-level summary across benchmarks and is reproduced without a subscript.
-- = not evaluated on that split.}
\label{tab:efficiency-std}
\end{table}

%% file: tables/table_difficulty_std.tex
\begin{table*}[!t]
\centering
\small
\setlength{\tabcolsep}{3pt}
\renewcommand{\arraystretch}{1.15}
\resizebox{\textwidth}{!}{%
\begin{tabular}{l!{\vrule}cc@{\hspace{7pt}}cc@{\hspace{7pt}}cc!{\vrule}cc@{\hspace{7pt}}cc@{\hspace{7pt}}cc}
\toprule
\multirow{3}{*}{\textbf{Setting}}
& \multicolumn{6}{c!{\vrule}}{\textbf{AppWorld}$_\text{TN}$}
& \multicolumn{6}{c}{\textbf{AppWorld}$_\text{TC}$} \\
\cmidrule(lr){2-7} \cmidrule(lr){8-13}
& \multicolumn{2}{c}{\textbf{Easy} (57)}
& \multicolumn{2}{c}{\textbf{Medium} (48)}
& \multicolumn{2}{c!{\vrule}}{\textbf{Hard} (63)}
& \multicolumn{2}{c}{\textbf{Easy} (72)}
& \multicolumn{2}{c}{\textbf{Medium} (150)}
& \multicolumn{2}{c}{\textbf{Hard} (195)} \\
\cmidrule(lr){2-3} \cmidrule(lr){4-5} \cmidrule(lr){6-7}
\cmidrule(lr){8-9} \cmidrule(lr){10-11} \cmidrule(lr){12-13}
& \textbf{TGC} & \textbf{SGC} & \textbf{TGC} & \textbf{SGC} & \textbf{TGC}
& \multicolumn{1}{c!{\vrule}}{\textbf{SGC}}
& \textbf{TGC} & \textbf{SGC} & \textbf{TGC} & \textbf{SGC} & \textbf{TGC} & \textbf{SGC} \\
\midrule
\rowcolor{blockhl}
\multicolumn{13}{@{}l}{\texttt{Qwen2.5-32B-Instruct}} \\
Base & 64.3$_{\pm 8.8}$ & 38.6$_{\pm 13.2}$ & 32.6$_{\pm 4.8}$ & 8.3$_{\pm 9.5}$ & 12.2$_{\pm 3.3}$ & 4.8$_{\pm 0.0}$ & 50.5$_{\pm 2.9}$ & 23.6$_{\pm 2.4}$ & 15.1$_{\pm 0.8}$ & 4.0$_{\pm 2.0}$ & 6.7$_{\pm 0.0}$ & 0.5$_{\pm 0.9}$ \\
\rowcolor{ourshl}
DRACO (Ours) & 83.2$_{\pm 2.9}$ & 66.7$_{\pm 5.3}$ & 66.9$_{\pm 7.6}$ & 44.4$_{\pm 11.5}$ & 41.4$_{\pm 2.1}$ & 18.5$_{\pm 8.0}$ & 77.5$_{\pm 1.4}$ & 57.4$_{\pm 5.3}$ & 22.6$_{\pm 2.8}$ & 9.3$_{\pm 1.3}$ & 26.7$_{\pm 0.9}$ & 12.7$_{\pm 2.6}$ \\
\midrule
\rowcolor{blockhl}
\multicolumn{13}{@{}l}{\texttt{Qwen3.6-27B}} \\
Base & 84.2$_{\pm 5.3}$ & 66.7$_{\pm 11.0}$ & 63.2$_{\pm 6.7}$ & 29.2$_{\pm 9.5}$ & 60.8$_{\pm 7.8}$ & 27.0$_{\pm 14.5}$ & 73.1$_{\pm 2.1}$ & 55.6$_{\pm 6.4}$ & 42.0$_{\pm 4.1}$ & 27.3$_{\pm 3.1}$ & 47.0$_{\pm 1.3}$ & 23.1$_{\pm 4.1}$ \\
\rowcolor{verifhl}
Outcome reward & 91.6$_{\pm 0.3}$ & 82.5$_{\pm 1.8}$ & 80.5$_{\pm 3.7}$ & 54.8$_{\pm 4.8}$ & 69.1$_{\pm 3.6}$ & 41.8$_{\pm 5.6}$ & 83.8$_{\pm 1.2}$ & 68.1$_{\pm 1.4}$ & 51.4$_{\pm 0.1}$ & 33.1$_{\pm 0.4}$ & 57.6$_{\pm 2.6}$ & 31.3$_{\pm 5.1}$ \\
DRACO w/o Dyn.\ \& Cred. & 91.4$_{\pm 1.2}$ & 81.3$_{\pm 1.0}$ & 81.0$_{\pm 2.4}$ & 53.5$_{\pm 7.3}$ & 71.8$_{\pm 3.2}$ & 45.5$_{\pm 8.7}$ & 78.2$_{\pm 2.1}$ & 64.8$_{\pm 4.5}$ & 51.4$_{\pm 1.4}$ & \textbf{36.7}$_{\pm 1.3}$ & 60.9$_{\pm 0.3}$ & 38.5$_{\pm 1.5}$ \\
DRACO w/o Dyn. & 91.8$_{\pm 3.8}$ & 80.1$_{\pm 10.0}$ & 84.7$_{\pm 1.2}$ & 60.4$_{\pm 5.5}$ & 70.9$_{\pm 3.2}$ & 43.4$_{\pm 1.8}$ & 77.3$_{\pm 0.5}$ & 61.1$_{\pm 3.7}$ & 51.3$_{\pm 2.1}$ & 34.4$_{\pm 3.3}$ & 59.1$_{\pm 1.0}$ & 34.9$_{\pm 1.8}$ \\
DRACO w/o Cred. & 92.0$_{\pm 1.8}$ & 81.3$_{\pm 3.7}$ & 81.2$_{\pm 2.1}$ & 58.3$_{\pm 4.2}$ & 73.7$_{\pm 1.2}$ & \textbf{55.0}$_{\pm 3.3}$ & 77.9$_{\pm 3.7}$ & 61.6$_{\pm 2.1}$ & 48.2$_{\pm 1.0}$ & 32.9$_{\pm 2.3}$ & 61.0$_{\pm 0.5}$ & 38.8$_{\pm 1.6}$ \\
\rowcolor{ourshl}
DRACO (Ours) & \textbf{93.4}$_{\pm 1.2}$ & \textbf{89.5}$_{\pm 3.0}$ & \textbf{88.4}$_{\pm 1.4}$ & \textbf{72.9}$_{\pm 4.2}$ & \textbf{75.7}$_{\pm 3.2}$ & 51.9$_{\pm 4.6}$ & \textbf{81.6}$_{\pm 2.5}$ & \textbf{65.3}$_{\pm 5.6}$ & 50.6$_{\pm 2.8}$ & 29.6$_{\pm 3.2}$ & \textbf{62.4}$_{\pm 0.6}$ & \textbf{40.2}$_{\pm 0.6}$ \\
\rowcolor{ourshl}
DRACO (Ours - self-judge) & 90.3$_{\pm 3.8}$ & 79.5$_{\pm 3.7}$ & 84.5$_{\pm 4.9}$ & 62.5$_{\pm 6.3}$ & 70.2$_{\pm 1.3}$ & 47.6$_{\pm 2.8}$ & 77.9$_{\pm 3.3}$ & 53.2$_{\pm 3.5}$ & \textbf{52.8}$_{\pm 1.6}$ & 27.6$_{\pm 2.8}$ & 61.1$_{\pm 1.4}$ & 33.3$_{\pm 3.6}$ \\
\bottomrule
\end{tabular}}
\caption{\textbf{Task-success by difficulty} on both AppWorld test splits, in \%, by the
AppWorld metadata label and scored against each stratum's own denominator (counts in the
header). Entries are the $p^1$ level of Table~\ref{tab:main} and follow its conventions
throughout. For trained rows, subscripts give the standard deviation over the three
checkpoints each mean averages, with three evaluation runs per checkpoint. Base rows are
single-checkpoint, so their subscripts instead give the standard deviation over three
independent evaluation runs.
-- = not evaluated on that split.}
\label{tab:difficulty-std}
\end{table*}

%% file: tables/table_selfjudge.tex
\begin{table*}[!t]
\centering
\small
\setlength{\tabcolsep}{4pt}
\renewcommand{\arraystretch}{1.15}
\resizebox{\textwidth}{!}{%
\begin{tabular}{l!{\vrule}ccc@{\hspace{8pt}}ccc!{\vrule}ccc@{\hspace{8pt}}ccc!{\vrule}ccc!{\vrule}ccc}
\toprule
\multirow{3}{*}{\textbf{Setting}}
& \multicolumn{6}{c!{\vrule}}{\textbf{AppWorld}$_\text{TN}$}
& \multicolumn{6}{c!{\vrule}}{\textbf{AppWorld}$_\text{TC}$}
& \multicolumn{3}{c!{\vrule}}{$\boldsymbol{\tau}_\text{B}$}
& \multicolumn{3}{c}{} \\
\cmidrule(lr){2-7} \cmidrule(lr){8-13} \cmidrule(lr){14-16}
& \multicolumn{3}{c}{\textbf{TGC}} & \multicolumn{3}{c!{\vrule}}{\textbf{SGC}}
& \multicolumn{3}{c}{\textbf{TGC}} & \multicolumn{3}{c!{\vrule}}{\textbf{SGC}}
& \multicolumn{3}{c!{\vrule}}{\textbf{SR}}
& \multicolumn{3}{c}{\multirow{-2}{*}{\textbf{Average}}} \\
\cmidrule(lr){2-4} \cmidrule(lr){5-7} \cmidrule(lr){8-10} \cmidrule(lr){11-13} \cmidrule(lr){14-16} \cmidrule(lr){17-19}
& $p^1$ & $p^2$ & $p^3$ & $p^1$ & $p^2$ & \multicolumn{1}{c!{\vrule}}{$p^3$}
& $p^1$ & $p^2$ & $p^3$ & $p^1$ & $p^2$ & \multicolumn{1}{c!{\vrule}}{$p^3$}
& $p^1$ & $p^2$ & \multicolumn{1}{c!{\vrule}}{$p^3$}
& $p^1$ & $p^2$ & $p^3$ \\
\midrule
\rowcolor{ourshl}
DRACO (self-judge) & 81.1 & 72.1 & 65.5 & 62.7 & 48.2 & 38.1 & 61.0 & 45.6 & 36.9 & 34.7 & 21.7 & 16.1 & 21.1 & 14.7 & 11.7 & 52.1 & 40.5 & 33.7 \\
DRACO (self-judge w/o Think.\ \& $k{=}3$) & 79.3 & 68.2 & 60.9 & 59.3 & 45.6 & 38.1 & 54.3 & 43.0 & 36.7 & 36.2 & 25.0 & 18.0 & 18.6 & 11.7 & 9.6 & 49.5 & 38.7 & 32.7 \\
\bottomrule
\end{tabular}}
\caption{\textbf{Self-judge variants} on \texttt{Qwen3.6-27B}, following the protocol and
conventions of Table~\ref{tab:main}. The first row is the self-judge of the main text:
thinking enabled in every judge phase, and scoring repeated $k{=}3$ times per trajectory
with a criterion counted as passed only if all three calls pass it. The second row
disables thinking and scores with a single call.}
\label{tab:selfjudge-variants}
\end{table*}

%% file: tables/table_frontier.tex
\begin{table}[!t]
\centering
\small
\setlength{\tabcolsep}{4pt}
\renewcommand{\arraystretch}{1.15}
\resizebox{\columnwidth}{!}{%
\begin{tabular}{@{}l!{\vrule}cc!{\vrule}cc@{}}
\toprule
\textbf{Model} & \textbf{TGC} & \textbf{SGC} & \textbf{Cost (\$)} & \textbf{\$/task} \\
\midrule
\texttt{claude-opus-4-7}        & \textbf{93.5} & \textbf{91.1} & 32.71 & 0.195 \\
\texttt{claude-sonnet-4-6}      & 89.9 & 85.7 & 18.81 & 0.112 \\
\texttt{Kimi-K2.7-Code}         & 89.9 & 82.1 &  6.65 & 0.040 \\
\texttt{DeepSeek-V4-Flash}      & 70.8 & 50.0 &  2.42 & 0.014 \\
\texttt{gpt-oss-120b}           & 44.0 & 26.8 &  3.50 & 0.021 \\
\texttt{gemini-2.5-flash-lite}  & 33.9 & 12.5 &  \textbf{0.92} & \textbf{0.005} \\
\midrule
\texttt{Qwen3.6-27B} (untrained) & 69.4 & 41.1 & 10.77 & 0.064 \\
\rowcolor{ourshl}
DRACO (\texttt{Qwen3.6-27B})    & 85.3 & 70.6 &  8.27 & 0.049 \\
\bottomrule
\end{tabular}}
\caption{\textbf{Frontier models vs.\ DRACO} on \awtn (168 tasks), in \%, with the
US-dollar cost of one pass over the split and per task. Frontier models: one run each,
priced at the provider's API rates. Qwen rows: protocol and reference price of
Table~\ref{tab:main}. Best per column in bold.}
\label{tab:frontier}
\end{table}

%% file: tables/table_static_rubrics.tex
\begin{table*}[t]
\centering
\small
\setlength{\tabcolsep}{4pt}
\renewcommand{\arraystretch}{1.15}
\begin{tabular}{@{}rp{0.20\textwidth}p{0.62\textwidth}@{}}
\toprule
\# & Title & Description \\
\midrule
1 & Protects secret values & The agent should avoid printing or restating passwords, tokens, or other secrets. Using a secret inside an API call is fine, but exposing it is not. \\
2 & Stops after completion & After the task is completed, the agent should stop sending more code. Extra code after completion is unnecessary and risky. \\
3 & Fetches complete data & The agent should collect all needed records before deciding. This includes checking later pages, reading needed details, and not trusting visibly truncated output as complete. \\
4 & Completes task properly & The agent should explicitly finish with one completion call after the work is done. If an answer is required, the completion call should contain only the requested value or type. \\
5 & Computes from live data & The agent should work from returned objects and compute results in code when calculation is needed. Manual retyping, eyeballing, or unsupported assumptions should fail. \\
6 & Handles time correctly & For time-based tasks, the agent should get the current time from the environment and use the correct calendar boundaries. Hardcoded dates or wrong windows should fail. \\
7 & Processes items completely & The agent should cover the full required set of items and handle duplicates or repeated occurrences correctly. Missing items, double-counting, or acting only once when each occurrence matters should fail here. \\
8 & Targets exact items & When making changes, the agent should act on exactly the intended items and preserve any required exceptions. Wrong targets, overbroad edits, or changing protected items should fail here. \\
9 & Applies correct selection logic & The agent should identify targets using the exact task conditions. Wrong filters, wrong fields, wrong matching keys, or wrong comparison rules should fail here. \\
10 & Uses complete working set & After fetching data, the agent should use all relevant fetched items in later reasoning. Fetching extra pages or sources but then ignoring part of them is still incomplete. \\
11 & Performs required actions & The agent should actually carry out the requested create, update, delete, send, move, or similar actions. Listing targets without acting, or using the wrong action type, should fail. \\
12 & Verifies action outcome & After changing state, the agent should confirm that the action worked. A follow-up read or a clear success response can both count when they directly confirm the intended result. \\
13 & Uses exact action content & When an action needs exact text, names, amounts, paths, or preserved identity, the agent should use the exact required values. Unrequested renaming or altered content should fail. \\
14 & Uses correct data source & The agent should gather evidence from the app, endpoint, or object set the task actually requires. Using a different source that could change the result should fail. \\
15 & Authenticates correctly & For protected APIs, the agent should log in with the documented identity fields and then use the returned token. Missing login, wrong login fields, or missing token are distinct auth failures and should be merged here. \\
16 & Retrieves needed credentials & When login is required, the agent should get credentials from the supervisor data. Guessing, inventing, or hardcoding unseen secrets is not acceptable. \\
17 & Uses runnable Python & The agent should act through executable Python in the REPL. Non-code payloads or syntax-breaking text prevent normal execution. \\
18 & Checks docs when needed & The agent should consult app or API docs when it does not already know how to call something. Repeated blind guessing should fail this rubric. \\
19 & Recovers from failures & When code or API calls fail, the agent should change approach and continue productively. Repeating the same mistake, or endlessly retrying after a persistent failure pattern, should fail here. \\
20 & Uses valid API calls & The agent should use real documented endpoints with supported parameters. Made-up APIs, wrong endpoints, or unsupported arguments are the same observable mistake. \\
21 & Uses allowed tools only & The agent should stay within provided app APIs and allowed Python tools. Forbidden imports, OS access, or invented helpers are not allowed. \\
\bottomrule
\end{tabular}
\caption{\textbf{The static rubric set} (21 criteria), ordered by failure rate on the 720
base policy model rollouts used to author it. The \texttt{evaluator\_instruction} field is
omitted for space.}
\label{tab:staticrubrics}
\end{table*}